\documentclass[times,twocolumn,final]{elsarticle}

\usepackage{medima}
\usepackage{framed,multirow}
\usepackage{multicol}

\usepackage{amssymb}
\usepackage{latexsym}

\usepackage{url}
\usepackage{xcolor}

\usepackage{subfigure}
\usepackage{bm}
\usepackage{amsmath,amsfonts}
\usepackage{algorithmic}
\usepackage{graphicx}
\usepackage{textcomp}
\usepackage{subfigure}
\usepackage{mathrsfs}
\usepackage{pifont}
\usepackage{stfloats}
\usepackage{booktabs}

\usepackage[colorlinks,
linkcolor=blue,
anchorcolor=blue,
citecolor=blue,
urlcolor=blue]{hyperref}

\definecolor{newcolor}{rgb}{.8,.349,.1}

\journal{Medical Image Analysis}

\begin{document}

\verso{Yu Cai \textit{et~al.}}

\begin{frontmatter}

\title{Dual-distribution discrepancy with self-supervised refinement for anomaly detection in medical images}

\author[1]{Yu \snm{Cai}}
\author[2,4]{Hao \snm{Chen}\corref{cor1}}
\cortext[cor1]{Corresponding author.}
\ead{jhc@cse.ust.hk}
\author[3]{Xin \snm{Yang}}
\author[3]{Yu \snm{Zhou}}
\author[1,2]{Kwang-Ting \snm{Cheng}}

\address[1]{Department of Electronic and Computer Engineering, The Hong Kong University of Science and Technology, Hong Kong, China}
\address[2]{Department of Computer Science and Engineering, The Hong Kong University of Science and Technology, Hong Kong, China}
\address[4]{Department of Chemical and Biological Engineering, The Hong Kong University of Science and Technology, Hong Kong, China}
\address[3]{School of Electronic Information and Communications, Huazhong University of Science and Technology, Wuhan 430074, China}

\received{22 August 2022}
\finalform{13 February 2023}
\accepted{6 March 2023}
\availableonline{13 March 2023}

\begin{abstract}
Medical anomaly detection is a crucial yet challenging task aimed at recognizing abnormal images to assist in diagnosis. Due to the high-cost annotations of abnormal images, most methods utilize only known normal images during training and identify samples deviating from the normal profile as anomalies in the testing phase. Many readily available unlabeled images containing anomalies are thus ignored in the training phase, restricting the performance. To solve this problem, we introduce one-class semi-supervised learning (OC-SSL) to utilize known normal and unlabeled images for training, and propose Dual-distribution Discrepancy for Anomaly Detection (DDAD) based on this setting. Ensembles of reconstruction networks are designed to model the distribution of normal images and the distribution of both normal and unlabeled images, deriving the normative distribution module (NDM) and unknown distribution module (UDM). Subsequently, the intra-discrepancy of NDM and inter-discrepancy between the two modules are designed as anomaly scores. Furthermore, we propose a new perspective on self-supervised learning, which is designed to refine the anomaly scores rather than directly detect anomalies. Five medical datasets, including chest X-rays, brain MRIs and retinal fundus images, are organized as benchmarks for evaluation. Experiments on these benchmarks comprehensively compare a wide range of anomaly detection methods and demonstrate that our method achieves significant gains and outperforms the state-of-the-art. Code and organized benchmarks are available at \url{https://github.com/caiyu6666/DDAD-ASR}.
\end{abstract}


\end{frontmatter}


\section{Introduction}
\label{sec:introduction}
Medical imaging is of vital importance to the diagnosis of a wide variety of pathologies. Take the case of chest X-rays (CXRs), which are the most commonly performed radiological exam \citep{ccalli2021deep}, widely applied for the diagnosis of tens of lung diseases such as pneumonia, nodules, lung opacity, pneumothorax, etc. To alleviate the burden on radiologists in reading CXRs and improve diagnosis efficiency, automatic CXR analysis using deep learning is becoming popular \citep{luo2020deep,luo2021oxnet,luo2022rethinking,10.1007/978-3-031-16452-1_59}. However, such methods require the annotation of images, which is difficult, relies on the experience of professional experts, and is time-consuming and labor-intensive. This has motivated the development of intelligent systems to help radiologists automatically detect and localize potential abnormalities using few or even no annotations. 


\begin{figure}[!t]
\centering
\begin{minipage}[t]{1\linewidth}
\centering
\includegraphics[width=1\linewidth]{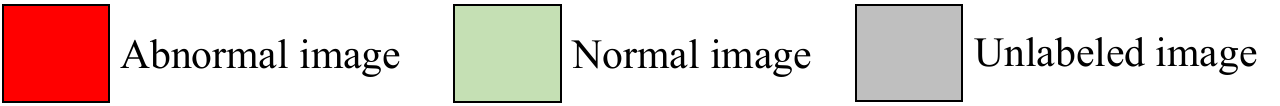}
\end{minipage}

\subfigure[One-class]{
\begin{minipage}[t]{0.28\linewidth}
\centering
\includegraphics[height=0.5in]{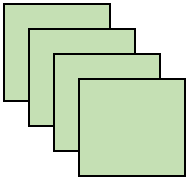}
\end{minipage}
\label{fig1:occ}
}
\subfigure[Semi-supervised]{
\begin{minipage}[t]{0.28\linewidth}
\centering
\includegraphics[height=0.5in]{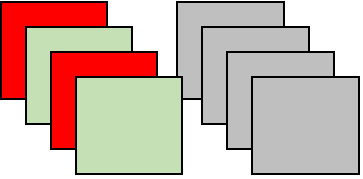}
\end{minipage}
\label{fig1:semi}
}
\subfigure[OC-SSL]{
\begin{minipage}[t]{0.28\linewidth}
\centering
\includegraphics[height=0.5in]{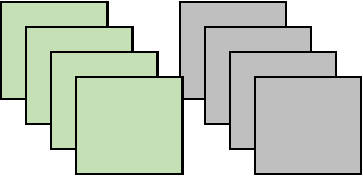}
\end{minipage}
\label{fig1:ours}
}
\caption{Different training modes for medical anomaly detection. (a) One-class classification mode, utilizing only normal images, is the most popular, but wastes unlabeled images. (b) Semi-supervised mode requires labeled normal and abnormal images, and mostly normal unlabeled images, thus infeasible in clinical practice. (c) The introduced OC-SSL mode utilizes normal and unlabeled images with arbitrary anomaly rates.}
\label{fig1}
\end{figure}

As annotations of normal images from healthy subjects are relatively easy to obtain while those of anomalies are complex, various and usually difficult to collect, most existing methods consider anomaly detection as a one-class classification (OCC) problem \citep{ruff2018deep}, where only normal images are utilized for training and samples not conforming to the normal profile are identified as anomalies in the testing phase; thus there is no need for annotation of abnormal images during training. This setting has been extensively studied in anomaly detection for both computer vision tasks \citep{ruff2021unifying} and medical image analysis \citep{baur2021autoencoders}. Nevertheless, due to the lack of training on real abnormal images, the discriminative capability of these methods is limited. Meanwhile, in medical images analysis, an important fact is ignored that, different from the application scenarios in computer vision tasks, like industrial defect detection \citep{bergmann2019mvtec} and video anomaly detection \citep{sultani2018real,li2013anomaly}, where abnormal cases are rare, medical clinical practice provides plenty of readily available unlabeled images with a certain anomaly rate (AR). These unlabeled images, containing rich anomalous features, are wasted by methods based on the OCC setting, which restricts the performance of anomaly detection. 

Although some works have explored the utilization of unlabeled samples, the unlabeled abnormal samples have yet to be exploited successfully. Deep SAD \citep{ruff2019deep} introduced semi-supervised anomaly detection, however, it works under the condition that both labeled normal and abnormal samples are available, while the unlabeled data is mostly normal. This condition is difficult to achieve in practice, while anomalies in unlabeled data are not exploited. One-class SVM (OC-SVM) \citep{scholkopf1999support} and Support Vector Data Description (SVDD) \citep{tax2004support} utilize nonzero slack variables to penalize the objective function and learn soft margins, and thus tolerate a small number of outliers in the training set. 
However, they essentially try to reduce the effects of unlabeled abnormal samples for training on normal data similar to the OCC setting, rather than capture useful information from the unlabeled abnormal samples. It has been demonstrated that their performance will decrease consistently as the abnormal samples in the unlabeled data increase \citep{yoon2022selfsupervise}. Up to now, there is still no notable work leveraging unlabeled images for anomaly detection effectively.

\begin{figure}[ht]
\centering
\subfigure[Standard self-supervised anomaly detection]{
\begin{minipage}[t]{\linewidth}
    \centering
    \includegraphics[width=\linewidth]{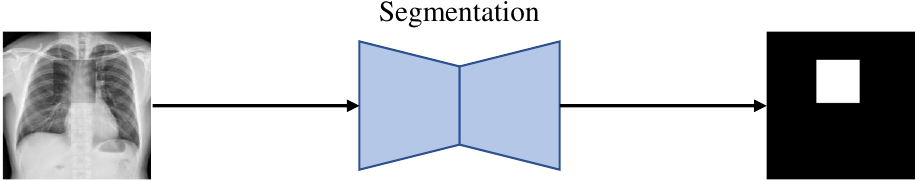}
\end{minipage}%
} \\
\subfigure[The proposed self-supervised anomaly score refinement]{
\begin{minipage}[t]{\linewidth}
    \centering
    \includegraphics[width=\linewidth]{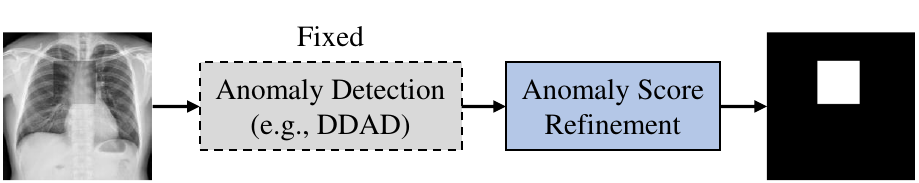}
\end{minipage}
}
\caption{Comparison of (a) the standard self-supervised anomaly detection and (b) the proposed self-supervised anomaly score refinement. (a) trains the network to directly detect the synthetic abnormal patterns from the input image, while (b) learns to refine the original anomaly score maps for the final accurate abnormal regions.}
\label{ASR}
\end{figure}

Therefore, a question is naturally raised: can unlabeled images provide effective information of abnormalities as a complement to normal images to improve the performance of anomaly detection? Motivated by this question, in this work, we introduce and explore one-class semi-supervised learning (OC-SSL) to train the model on known normal and unlabeled images. A comparison of the OC-SSL with existing settings is shown in Fig.~\ref{fig1}. As mentioned above, the OCC mode (Fig.~\ref{fig1:occ}) has been extensively studied in most existing anomaly detection works, but plenty of unlabeled images are ignored. Existing semi-supervised anomaly detection methods (Fig.~\ref{fig1:semi}) \citep{ruff2019deep} require both labeled normal and abnormal samples, while the unlabeled data should be mostly normal. It is intractable in practice, while unlabeled abnormal samples are not exploited. The introduced OC-SSL mode (Fig.~\ref{fig1:ours}) is capable of utilizing normal and unlabeled images with arbitrary ARs, while there is no need for labeled abnormal images. Therefore, the OC-SSL is more reasonable and consistent with the medical clinical practice.

\begin{figure*}
\centering
\includegraphics[width=1\linewidth]{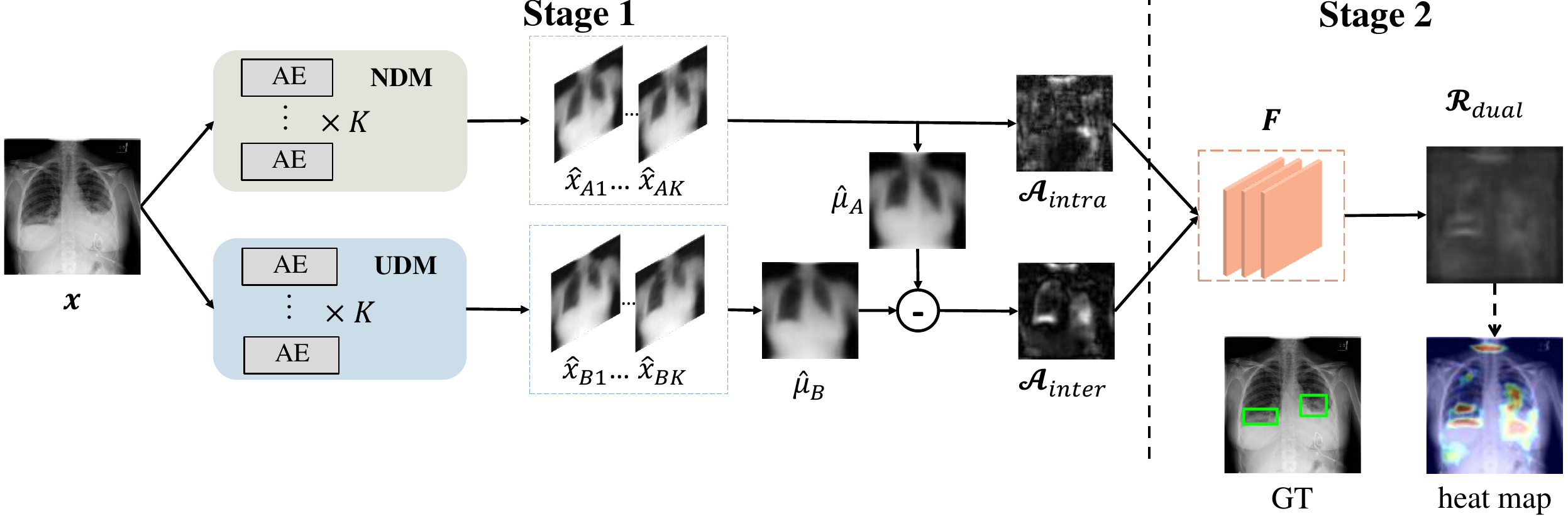}
\caption{Overview of the proposed DDAD. In the Stage 1, NDM and UDM model the distribution of known normal images and the distribution of known normal and unlabeled images, respectively. Then the intra-discrepancy inside NDM and inter-discrepancy between the two modules are designed as anomaly scores. In the Stage 2, the two anomaly scores are refined and fused by the ASR-Net $F(\cdot)$, deriving the final prediction $\mathcal{R}_{dual}$.}
\label{DDAD}
\end{figure*}

Based on the OC-SSL mode, we propose Dual-distribution Discrepancy for Anomaly Detection (DDAD), as shown in Fig.~\ref{DDAD}. To capture information from both known normal images and unlabeled images, we utilize ensembles of reconstruction networks to model the distribution of normal images and the distribution of both normal and unlabeled images, deriving the normative distribution module (NDM) and unknown distribution module (UDM). Subsequently, the intra-discrepancy of NDM and inter-discrepancy between the two modules are designed as anomaly scores (ASs). To further refine the two ASs, we design an Anomaly Score Refinement Net (ASR-Net), which is trained via self-supervised learning. Fig.~\ref{ASR} depicts our comparison with the standard self-supervised anomaly detection. Instead of learning to directly detect the synthetic abnormal patterns, the proposed ASR-Net learns to map the original AS to the final accurate abnormal regions, thereby avoiding overfitting and leading to better performance.
Considering the lack of publicly available benchmarks for medical anomaly detection, we for the first time collect and organize five medical datasets including CXRs, brain MRIs and retinal fundus images for evaluation and release them to facilitate other researchers evaluating their methods fairly. Experiments on these five datasets demonstrate that the proposed DDAD outperforms existing state-of-the-art methods, even if without unlabeled images, while unlabeled images can be utilized to further improve our performance by a large margin. Evaluation on unseen diseases further demonstrates the potential of our method for recognition of rare diseases, whose samples are inaccessible in the unlabeled data. A comprehensive comparison of a wide range of anomaly detection methods is also provided on the five datasets, revealing the performance of different families of methods and potential trends.

Our main contributions are summarized as follows:
\begin{itemize}
\item One-class semi-supervised learning (OC-SSL) is introduced. It utilizes known normal and unlabeled images with arbitrary ARs for anomaly detection, and is reasonable and consistent with clinical practice.

\item Based on the OC-SSL setting, ensembles of reconstruction networks are used to model the distribution of training data in an unsupervised fashion. Specifically, the NDM and UDM are designed to model the distribution of known normal images and the distribution of known normal and unlabeled images, respectively. It is the first time that unlabeled images are utilized to improve the performance of anomaly detection.

\item Two novel and powerful ASs, the \textit{intra-discrepancy} inside NDM and \textit{inter-discrepancy} between the NDM and UDM, are proposed to indicate anomalies.

\item An Anomaly Score Refinement Net (ASR-Net), trained via self-supervised learning, is proposed to refine and fuse the two ASs. Different from existing self-supervised anomaly detection methods that learn to detect synthetic abnormal patterns, it provides a new perspective on self-supervised learning, i.e., learning to map the original AS to the final accurate abnormal regions. It avoids overfitting and achieves better performance.

\item Five medical datasets that include three modalities are collected and organized, and released as benchmarks for medical anomaly detection. These facilitate a fair comparison with other methods as there are few related existing benchmarks.

\item Extensive experiments on the five medical datasets demonstrate that the proposed method achieves consistent, significant gains and outperforms state-of-the-art methods in anomaly detection. A comprehensive comparison of a wide range of anomaly detection methods is provided to reveal the performance of different families of methods and potential trends.
\end{itemize}

A preliminary version of this work was early accepted for MICCAI 2022 \citep{cai2022dual}. In this paper, the major extensions include designing a new module, namely ASR-Net, that provides a new perspective on self-supervised learning in anomaly detection and improves the performance and robustness significantly; adding much more experiments on more datasets containing different modalities; elaborating the analysis; and providing a more comprehensive literature review.

The rest of this paper is organized as follows: Section \ref{sec:related work} presents related works. Section \ref{sec:method} describes in detail the proposed DDAD methods with our ASR-Net. In Section \ref{Experiments}, extensive experiments on five datasets are conducted to demonstrate the effectiveness of our proposed method. Section \ref{sec:Discussion} discusses advantages and limitations of the proposed method, and analyzes a wide variety of methods to reveal future directions and trends. We conclude our work in Section \ref{sec:Conclusion}.

\section{Related works}
\label{sec:related work}
Anomaly detection aims at finding patterns in data that do not conform to expected behavior \citep{chandola2009anomaly}. It is a promising field that has been widely applied in a variety of domains. Due to the difficulty of collecting abundant annotated abnormal samples, almost all the existing works utilize only normal images during training, which is the well-known OCC setting \citep{ruff2018deep}.

Classical anomaly detection methods, OC-SVM \citep{scholkopf1999support} and SVDD \citep{tax2004support}, often fail in high-dimensional data due to bad computational scalability and the curse of dimensionality. Their derived Deep SVDD \citep{ruff2018deep} utilizes neural networks to constrain the normal samples in a hypersphere with minimum volume, handling high-dimensional data better but suffering from the mode collapse. Most recent state-of-the-art anomaly detection methods focus on reconstruction and self-supervised learning. As techniques highly related to our work, ensemble-based uncertainty estimates and semi-supervised learning for anomaly detection are also described in this section.

\subsection{Reconstruction-based Anomaly Detection}
Reconstruction-based methods are one of the most popular families in anomaly detection, especially for medical images \citep{baur2021autoencoders}. They usually utilize generative models, such as generative adversarial networks (GANs) \citep{goodfellow2014generative}, auto-encoders (AEs) or their variants, to learn a mapping function to reconstruct normal images, while the unseen abnormal images are assumed unable to be reconstructed well by these models trained with only normal images, and in turn yield high reconstruction error. 

\citet{schlegl2017unsupervised} are the first to use GANs for anomaly detection. They proposed AnoGAN to learn the manifold of normal images. For a query image, a latent feature is found via an iterative process to generate an image most similar to the query image. The query image will be identified as abnormal if there is a large difference with the best generated image. To replace the time-consuming iterative process in the testing phase, \citet{schlegl2019f} further utilized an encoder to learn the mapping from the retinal OCT image to the latent space, and derived a fast version of AnoGAN, named f-AnoGAN. However, these GAN-based methods could suffer from memorization pitfalls, causing reconstructions to differ anatomically from the actual input.

Various approaches also used variants of AEs for anomaly detection, including Variational AE (VAE) \citep{zimmerer2018context}, Adversarial AE (AAE) \citep{chen2018unsupervised}, Vector Quantized VAE (VQ-VAE) \citep{marimont2021anomaly}, etc. To avoid abnormal images being well reconstructed, \citet{gong2019memorizing} proposed to augment the AE with a memory module, which can store the latent features of normal training samples. The reconstruction is obtained from a few most relevant memory records, thus tending to be close to a normal image and enlarging the reconstruction errors of abnormal images. Compared with GAN-based methods, AE-based methods can preserve more anatomical coherence, but usually generate blurry reconstructions \citep{baur2021autoencoders}, leading to false positive detection around high-frequency regions (e.g., boundaries). To mitigate this problem, \citet{mao2020abnormality} proposed to automatically estimate the pixel-level uncertainty of reconstruction using an AE, which is used to normalize the reconstruction error and suppress the false positive detection in CXRs significantly. 

Recently, incorporating adversarial training into AEs has become popular, as it combines the advantages of both. \citet{baur2018deep} demonstrated that AEs with spatial bottlenecks can reconstruct important fine details better than those with dense bottlenecks, and combined the spatial VAE with GAN to improve the realism of reconstructed normal samples for anomaly detection in brain MRIs. In addition to adversarial training, \citet{akcay2018ganomaly} used an extra encoder to map the reconstructed image to the latent space again, and minimized reconstruction errors in both the image space and latent space during training to aid in learning the data distribution for the normal samples. \citet{zaheer2020old} proposed to transform the fundamental role of a discriminator from identifying real and fake data to distinguishing between good and bad quality reconstructions, which is highly desirable in anomaly detection as a trained AE would not produce as good reconstructions for abnormal images as they would for normal images conforming to the learned representations.

\subsection{Self-Supervised Learning-based Anomaly Detection}
Self-supervised learning \citep{jing2020self}, referring to learning methods in which networks are explicitly trained using pretext tasks with generated pseudo labels, has also been extensively studied for anomaly detection. \citet{sohn2020learning} proposed to first learn self-supervised representations from one-class data and then build one-class classifiers on learned representations. Based on their proposed framework, they applied distribution augmentation \citep{jun2020distribution} for one-class contrastive learning to reduce the uniformity of representations. Further, \citet{tian2021constrained} combined distribution-augmented contrastive learning \citep{sohn2020learning}, augmentation prediction \citep{golan2018deep}, and position prediction \citep{doersch2015unsupervised} to learn feature representations for anomaly-sensitive detection models. Moreover, \citet{li2021cutpaste} proposed to learn representations by classifying normal data from their designed CutPaste, and then build a Gaussian density estimator on learned representations.

In addition to the aforementioned representation-based methods, some works \citep{tan2020detecting,tan2021detecting,schluter2022natural} proposed to manually synthesize defects to train models to detect irregularities. Various image processing approaches have been designed to synthesize abnormal images, including CutPaste \citep{li2021cutpaste}, Foreign Patch Interpolation (FPI) \citep{tan2020detecting}, Poisson Image Interpolation (PII) \citep{tan2021detecting}, etc. Recently, \citet{schluter2022natural} integrated Poisson image editing with rescaling, shifting and a new Gamma-distribution-based patch shape sampling strategy to synthesize natural and diverse anomalies. Background constraints and pixel-level labels derived from the resulting difference to the normal image were designed to make the results more relevant to the task.
However, these methods may not generalize well due to the inherent reliance on the similarity between synthetic abnormal patterns and real anomalies.

Also, \citet{zavrtanik2021draem} proposed to combine the reconstruction network with a self-supervised network. It feeds the concatenation of the original image and reconstruction result to a segmentation network trained via self-supervised learning, which is expected to learn a distance function between the original and reconstructed anomaly appearance. However, the self-supervised network could learn a shortcut to directly segment the synthesized anomalies, which is more accessible than learning the distance function. As a result, it still suffers from overfitting.

\subsection{Ensemble-based Uncertainty Estimates}
Deep Ensemble \citep{lakshminarayanan2017simple} is a simple but effective method for uncertainty estimates of deep neural networks, where high uncertainty will be expressed on out-of-distribution (OOD) samples. It has been successfully applied in the fields of open-set recognition and active learning \citep{beluch2018power}. However, supervised training, like semantic segmentation or classification, is required in these methods, which is always undesirable in anomaly detection. 

Recently, \citet{bergmann2020uninformed} proposed to utilize feature vectors of pretrained networks on normal regions as surrogate labels for the training of an ensemble of student networks, whose predictive variance was used as an AS to segment anomalous regions. They designed the ensemble-based method for industrial anomaly detection with no demand for labels, but required a powerful pretrained model, such as networks trained on ImageNet \citep{krizhevsky2012imagenet}.

\subsection{Semi-Supervised Learning for Anomaly Detection}
Semi-supervised learning \citep{chapelle2009semi} is a learning paradigm in which the algorithm is provided with some labeled samples as well as unlabeled samples to improve the performance. Due to the advantages of leveraging unlabeled data, it is especially widely used in medical image analysis, where annotations are expensive and the amount of unlabeled data is huge. However, semi-supervised learning has not been successfully employed for medical anomaly detection due to two challenges. The first is that in anomaly detection, only normal images comprise the labeled data, which is inadequate for existing semi-supervised methods. Secondly, there are thousands of rare diseases, meaning that even though the unlabeled data may contain some types of anomalies, the testing data may contain many unseen types. It has been demonstrated that this mismatch can cause drastic performance degradation in semi-supervised learning \citep{oliver2018realistic}. 

Several attempts have been made to study semi-supervised learning for anomaly detection, but the two challenges remain unresolved. \citet{bauman2018one} proposed a semi-supervised learning algorithm for one-class classification. However, their setting is essentially transductive learning, where the model is directly tested on the unlabeled set. This is undesirable as, in practice, the trained model needs to be capable of finding anomalies from new data. Recently, \citet{ruff2019deep} introduced Deep SAD for general semi-supervised anomaly detection. However, it works under the condition that there are a few labeled normal and abnormal samples, while the unlabeled data is mostly normal. This condition is difficult to achieve in practice, while anomalies in unlabeled data are not exploited. Some works \citep{akcay2018ganomaly} refer to methods that train on only normal samples as ``semi-supervised''. Considering that only normal data is used for training, they are more precisely instances of one-class classification. Therefore, how to design an effective semi-supervised method or a variant to exploit unlabeled data for anomaly detection is still under study.

\subsection{Summary}

In summary, most of the previous anomaly detection methods used only normal images for training. Thus, plenty of unlabeled images in clinical practice were ignored. Although several works have tried to explore semi-supervised learning for anomaly detection, they work under strict conditions which do not meet clinical needs, and meanwhile no useful information is mined from the unlabeled data. To solve this problem, we introduce OC-SSL to train the model on known normal and unlabeled images.
We design the NDM and UDM, which are ensembles of several reconstruction networks, to model the normative distribution of normal images and unknown distribution of known normal and unlabeled images. Then the \emph{intra-discrepancy} inside the NDM and \emph{inter-discrepancy} between the NDM and UDM are used as the AS. 

Compared with previous reconstruction-based methods \citep{baur2021autoencoders}, our scores are the discrepancy among the outputs of network ensembles, rather than the discrepancy between the input and output. Therefore, more information can be captured, while the high reconstruction errors in normal regions, caused by reconstruction ambiguity or memorization pitfalls, can be mitigated in some way. Compared with existing ensemble-based methods \citep{bergmann2020uninformed}, we innovatively use reconstruction networks as the basic models for ensemble. They can be trained in an unsupervised fashion based on the images themselves, i.e., reconstruction. Therefore, neither labels nor pretrained models are required, meaning our method can be applied in various scenarios more easily, including but not limited to medical anomaly detection. Compared with previous attempts related to semi-supervised learning for anomaly detection, our OC-SSL setting requires only known normal and unlabeled images with arbitrary ARs for training, which greatly meets clinical needs. Also, through computing the inter-discrepancy between NDM and UDM, the unlabeled data can help the recognition of seen anomalies while no harm is caused to unseen anomalies, and thereby no performance degradation is caused by class distribution mismatch in the unlabeled data \citep{oliver2018realistic}.

We further propose ASR-Net trained via self-supervised learning to refine and fuse the two designed ASs. Different from existing self-supervised anomaly detection methods that require realistic pseudo abnormal images, it learns to map the original AS to the final accurate abnormal regions, and is thus insensitive to the synthetic abnormal images, yielding better generalization.

\section{Method}
\label{sec:method}

\subsection{Problem Statement}
In this section, we will first formulate the anomaly detection problem. The differences between our setting and the previous setting will also be clarified.  

Most existing works formulate anomaly detection as an OCC problem. That is, given a normal dataset $D_n=\{\bm{x}_{ni}\}_{i=1}^{N}$ with \emph{N} normal images, and a test dataset $D_t=\{(\bm{x}_{ti}, y_i)\}_{i=1}^{T}$ with \emph{T} annotated normal or abnormal images, where $y_i \in \{0, 1\}$ is the image label (0 for normal image and 1 for abnormal image), the goal is to train a model based on the normal image set $D_n$, which can identify anomalies in the test dataset $D_t$ during inference. 

Different from previous works, our proposed DDAD, based on the OC-SSL setting,  makes full use of the unlabeled images in clinical practice. Specifically, in addition to the normal dataset $D_n$, we also utilize a readily available unlabeled dataset $D_u=\{\bm{x}_{ui}\}_{i=1}^{M}$ with \emph{M} unlabeled images that includes both normal and abnormal images to improve the performance of anomaly detection.

\subsection{Dual-distribution Modeling}
\begin{figure}[!t]
    \centering
    \includegraphics[width=1\linewidth]{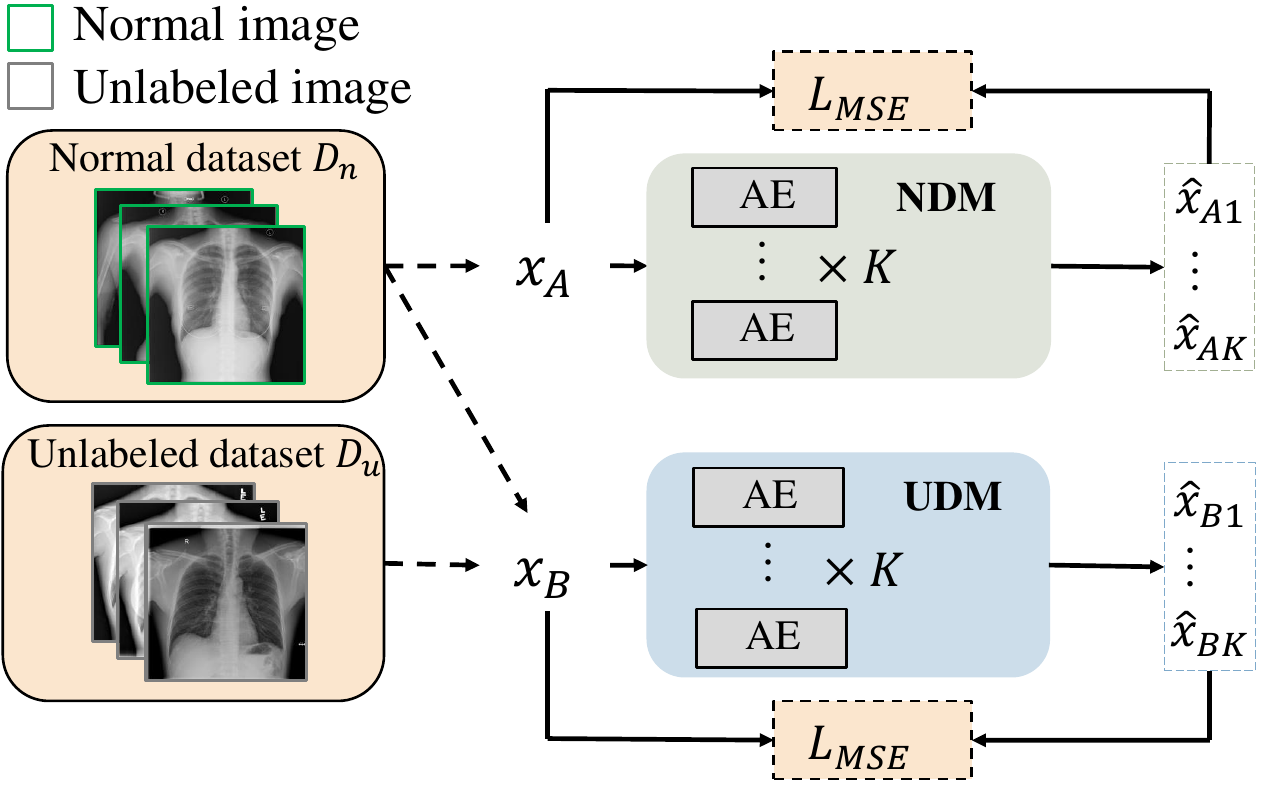}
    \caption{Illustration of training NDM and UDM.}
    \label{DDM}
\end{figure}

As shown in Fig.~\ref{DDAD}, we use two modules, the NDM and UDM, in Stage 1 to model the dual-distribution. The training process is illustrated in Fig.~\ref{DDM}. Each module is an ensemble of $K$ reconstruction networks with the same architecture but different random initialization of parameters and random shuffling of training samples, and is trained by the mean squared error (MSE) loss to minimize reconstruction errors on the training set. Specifically, the NDM is trained on only the normal dataset $D_n$ as
\begin{equation}
\mathcal{L}_{NDM} = \frac{1}{N} \sum_{\bm{x}_A \in D_n} \sum_{i=1}^{K} \Vert\bm{x}_A-\bm{\hat{x}}_{Ai}\Vert ^2,
\label{eq1}
\end{equation}
\noindent
where \emph{N} is the size of the normal dataset $D_n$, $\bm{x}_A$ is the input training image of the NDM, and $\bm{\hat{x}}_{Ai}$ is the reconstruction of $\bm{x}_A$ from the $i$-th network in the NDM.
Similarly, the loss function of UDM trained on both the normal image dataset $D_n$ and unlabeled dataset $D_u$ can be written as
\begin{equation}
\mathcal{L}_{UDM} = \frac{1}{N+M} \sum_{\bm{x}_B \in D_n \cup D_u} \sum_{i=1}^{K} \Vert\bm{x}_B-\bm{\hat{x}}_{Bi}\Vert ^2.
\label{eq2}
\end{equation}
\noindent
In this way, the NDM models the distribution of known normal images, while the UDM captures effective information of abnormalities from the unlabeled dataset as a complement to the normal images.

\subsection{Dual-distribution Discrepancy-based Anomaly Scores} \label{AS}
Given a testing image, the pixel-wise reconstruction error has been widely used as the anomaly score (AS). In this work, we design two innovative and effective ASs based on the proposed ensemble modules. 

Previous ensemble-based methods train the ensemble networks via supervised tasks like classification or regression, then utilize their output variance to identify OOD samples \citep{lakshminarayanan2017simple,bergmann2020uninformed}. In our DDAD, reconstruction networks are regarded as regressors that regress the gray value at each pixel. Therefore, based on Deep Ensemble \citep{lakshminarayanan2017simple}, the reconstructions' standard deviation can be used to estimate the samples' uncertainty. Specifically, as the networks in NDM are trained on only normal images, they will express a high difference on their OOD samples, i.e., abnormal regions. We propose to use this \emph{intra-discrepancy} inside the NDM as an AS:
\begin{equation}
\mathcal{A}_{intra}^p = \sqrt{\frac{1}{K} \sum_{i=1}^{K} (\hat{\mu}_A^p - \hat{x}_{Ai}^p)^2},
\end{equation}
\noindent
where $p$ is the index of pixels and $\hat{\mu}_A=\frac{1}{K} \sum_{i=1}^{K} \hat{x}_{Ai}$ is the average map of reconstructions from NDM. Meanwhile, as the UDM captures some anomalous features from unlabeled images that the NDM never sees, a high discrepancy between their outputs will also appear in these abnormal regions. We further propose to use this \emph{inter-discrepancy} between the two modules as another AS:
\begin{equation}
    \mathcal{A}_{inter}^p = \vert \hat{\mu}_A^p - \hat{\mu}_B^p \vert,
\end{equation}
\noindent
where $\hat{\mu}_B=\frac{1}{K} \sum_{i=1}^{K} \hat{x}_{Bi}$ is the average map of reconstructions from the UDM. As shown in Fig.~\ref{DDAD}, our discrepancy maps can indicate potential abnormal regions based on the pixel-wise AS. The image-level AS is obtained by averaging the pixel-level scores in each image.

Compared with $\mathcal{A}_{rec}$, our ASs consider the discrepancy between different distributions, leading to stronger discriminative capability. To understand why $\mathcal{A}_{inter}$ works, we can consider three situations: (1) When the testing input is a normal image, the NDM and UDM will have consistent reconstructions as they are both well-trained to reconstruct it, resulting in a small inter-discrepancy. (2) When the testing input is an abnormal image containing a disease that appears in the unlabeled dataset, the UDM will tend to have a different reconstruction to the NDM as the UDM has been trained to reconstruct this type of anomalies that the NDM never sees, leading to a high inter-discrepancy. (3) When the testing input is an abnormal image containing only diseases that never appear in the unlabeled dataset, it can be considered an OOD sample of the NDM and UDM; therefore, the $\mathcal{A}_{inter}$ performs similarly to the $\mathcal{A}_{intra}$ for this case.

Intuitively, seen diseases (situation (2)) can be distinguished better than unseen diseases (situation (3)) as the UDM has captured their information. Based on this hypothesis, a higher AR in the unlabeled data will increase seen abnormal samples and lead to a more competitive $\mathcal{A}_{inter}$. Therefore, our method is able to improve the performance on seen anomalies, while no harm is caused to unseen anomalies, i.e., no performance degradation caused by class distribution mismatch \citep{oliver2018realistic}.
Experiments in Section \ref{sec:exp_ablation} validate this hypothesis. In addition, the proposed method can achieve a consistent improvement compared with the reconstruction baseline even if the AR is 0, while a low AR can lead to a significant boost in performance. 

Our discrepancies are also all computed among reconstructions, rather than between the input and reconstruction as with $\mathcal{A}_{rec}$. This can reduce the false positive detection caused by the reconstruction ambiguity of the AE around high-frequency regions \citep{baur2021autoencoders,mao2020abnormality}.

\subsection{Uncertainty-refined Anomaly Scores}
Due to the reconstruction ambiguity of the AE, high reconstruction errors often appear at high-frequency regions, e.g., around normal region boundaries, leading to false positives. To address this problem, AE-U \citep{mao2020abnormality} was proposed to refine the $\mathcal{A}_{rec}$ using the estimated pixel-wise uncertainty. It generates the reconstruction $\hat{\bm{x}}_i$ and corresponding uncertainty $\bm{\sigma}^2(\bm{x}_i)$ for each input $\bm{x}_i$, and is trained by

\begin{equation}
\mathcal{L} = \frac{1}{NP} \sum_{i=1}^{N} \sum_{p=1}^{P} \{ \frac{(x_i^p-\hat{x}_i^p)^2}{\sigma_p^2(\bm{x}_i)} + {\rm log} \sigma_p^2(\bm{x}_i) \}.
\end{equation}

Training on normal images, the numerator of the first term is an \emph{MSE} loss to minimize the reconstruction error, while the $\sigma_p^2(\bm{x}_i)$ at the denominator will be learned automatically to be large at pixels with high reconstruction errors to minimize the first term. Additionally, the second term drives the predicted uncertainty to be small in other regions. The two loss terms together ensure that the predicted uncertainty will be larger at only normal regions with high reconstruction errors. Thus, it can be used to refine the AS at the pixel level.

In this work, we design a strategy similar to that of AE-U while adapting it to DDAD well. We use AE-U as the backbone of DDAD, and utilize the uncertainty predicted by our NDM, which is trained on only the normal dataset, to refine our intra- and inter-discrepancy at the $p$-th pixel as follows:
\begin{equation}
\mathcal{A}_{intra}^p = \frac{\sqrt{\frac{1}{K} \sum_{i=1}^{K} (\hat{\mu}_A^p - \hat{x}_{Ai}^p) ^2}}{\sigma_p},
\end{equation}
\begin{equation}
\mathcal{A}_{inter}^p = \frac{\vert \hat{\mu}_A^p - \hat{\mu}_B^p \vert}{\sigma_p},
\end{equation}
where $\sigma_p$ is the average uncertainty predicted by AE-U in the NDM.

\subsection{Self-supervised Learning-based Anomaly Score Refinement Net}

As shown in Fig.~\ref{DDAD}, the proposed $\mathcal{A}_{intra}$ and $\mathcal{A}_{inter}$ can overall express high values on abnormal regions, but some false positives and false negatives still appear. Based on the observations, we hypothesize that score maps can provide not only score values, but also spatial information to assist in the recognition of true positives. For example, false positives could be found around boundaries or noisy pixels. In this case, the discrepancy map on these regions would show the patterns as thin bright lines or small bright points, which are different from the patterns on real abnormal regions. Similarly, although the discrepancy value is low on false negatives, it could have some spatial patterns that are different from those of real normal regions. Therefore, we argue that false positive and false negative patterns in the score map can be recognized, based on which the score map can be further refined by eliminating false positives and recovering false negatives. To validate this hypothesis, we design an ASR-Net, denoted as $F(\cdot)$, to capture the spatial information in the raw discrepancy maps and refine them accordingly. Specifically, the network can be formulated as
\begin{equation}
\mathcal{R}_{dual} = F([\mathcal{A}_{intra}, \mathcal{A}_{inter}]),
\end{equation}
where the network $F(\cdot)$ takes the original dual-distribution discrepancy maps, $\mathcal{A}_{intra}$ and $\mathcal{A}_{inter}$, as inputs, and then predicts the final accurate AS map $\mathcal{R}_{dual}$ accordingly.

\begin{figure}[!t]
\centering
\includegraphics[width=1\linewidth]{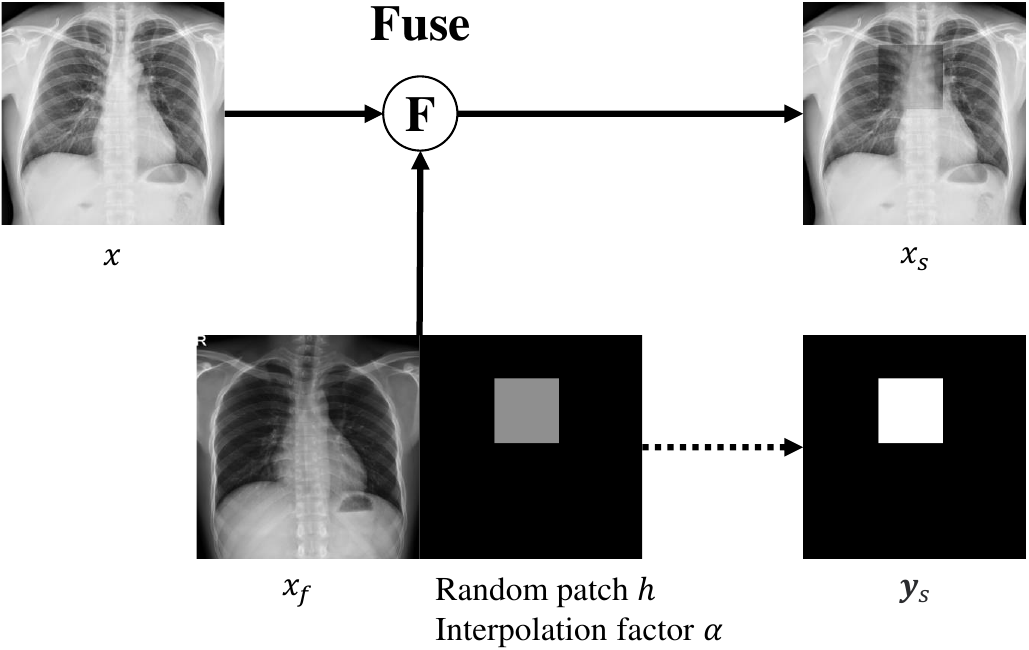}
\caption{Illustration of the synthesis of abnormal images. $x$ and $x_f$ denote two normal images, $x_s$ denotes the synthetic abnormal image and $\textbf{y}_s$ is the corresponding binary pseudo label.}
\label{synthesis}
\end{figure}

To obtain an effective $F(\cdot)$, we design a self-supervised task, where pseudo abnormal images with the corresponding pixel-level binary labels are synthesized to train $F(\cdot)$. Specifically, we employ a simple approach for the synthesis of abnormal images referenced to FPI \citep{tan2020detecting}. As shown in Fig.~\ref{synthesis}, for each normal image $x$, we assign a random patch $h$ and fuse $x$ with another normal image $x_f$ in the region $h$ with the interpolation $\alpha$, deriving synthetic abnormal image $x_s$. The operation is formulated as
\begin{equation}
x_s^p = (1 - \alpha) x^p + \alpha x_f^p, \forall{p \in h},
\end{equation}
where $p$ is the index of pixels and the interpolation $\alpha \sim U(0, 1)$. The random patch $h$ is restricted by:
\begin{equation}
    h_c \sim U(0.1d, 0.9d),
    h_s \sim U(0.1d, 0.4d),
\end{equation}
where $d$ is the image width, $h_c$ is the patch center coordinate and $h_s$ is the patch size.

After obtaining the synthetic abnormal image ${\bm x}_s$, we feed it through our well-trained NDM and UDM (i.e., Stage 1 in Fig.~\ref{DDAD}), and compute its $\mathcal{A}_{intra}$ and $\mathcal{A}_{inter}$. With the supervision of corresponding pseudo label ${\bm y}_s$, $F(\cdot)$ is then trained by the Focal Loss \citep{lin2017focal} as
\begin{equation}
    \mathcal{L}_{R} = \mathrm{FL}(F([\mathcal{A}_{intra}, \mathcal{A}_{inter}]), {\bm y}_s),
\end{equation}
where FL($\cdot$) is the Focal Loss function. For each pixel with prediction probability $p_t$ for the ground truth class, the focal loss is computed as
\begin{equation}
    \mathcal{L}_{focal}(p_t) = -(1-p_t)^\gamma \log(p_t),
\end{equation}
where $\gamma$ is the tunable focusing parameter.

In this way, the ASR-Net $F(\cdot)$ can automatically learn to predict final accurate abnormal regions based on the patterns in original score maps, as shown in Stage 2 of Fig.~\ref{DDAD}.
Different from previous self-supervised anomaly detection methods, ASR-Net learns the mapping function from the raw score maps to the final accurate abnormal regions, rather than learns to detect the synthetic abnormal patterns, achieving better generalization and less sensitivity to the quality of synthetic images.

In addition, for the case that the unlabeled images are not acquired, we also explore the performance of using only $\mathcal{A}_{intra}$ under the same setting as the OCC problem. The score map predicted by $F(\cdot)$ according to only $\mathcal{A}_{intra}$ is denoted as $\mathcal{R}_{intra}$:
\begin{equation}
    \mathcal{R}_{intra} = F(\mathcal{A}_{intra}).
\end{equation}

\section{Experiments}
\label{Experiments}

\begin{table*}[!t]
\centering
\caption{Summary of dataset repartitions. Note that $D_u$ is built using data selected from the images presented in parentheses without the use of their annotations.}
\label{data_repartition}
\resizebox{\linewidth}{!}{
\begin{tabular}{llll}
\hline
\multicolumn{1}{c}{\multirow{2}{*}{\textbf{Dataset}}} & \multicolumn{3}{c}{\textbf{Repartition}}                                              \\ \cline{2-4} 
\multicolumn{1}{c}{} & Normal Dataset $D_n$ & Unlabeled Dataset $D_u$                   & Testing Dataset $D_t$              \\ \hline
RSNA\textsuperscript{\ref{1}}                 & 3851                 & 4000 (4000 normal + 5012 abnormal images) & 1000 normal + 1000 abnormal images \\
VinDr-CXR\textsuperscript{\ref{2}} \citep{nguyen2022vindr}                                            & 4000 & 4000 (5606 normal + 3394 abnormal images) & 1000 normal + 1000 abnormal images \\
CXAD                 & 2000                 & 2000 (800 normal + 1200 abnormal images)  & 499 normal + 501 abnormal images   \\
Brain Tumor\textsuperscript{\ref{3}}             & 1000                 & 1000 (400 normal + 2666 abnormal images)  & 600 normal + 600 abnormal images   \\
LAG \citep{li2019attention}                  & 1500                 & 1500 (832 normal + 900 abnormal images)   & 811 normal + 811 abnormal images   \\ \hline
\end{tabular}
}
\end{table*}

\subsection{Datasets}
We conduct extensive experiments on three CXR datasets, one brain MRI dataset, and one retinal fundus image dataset: 1) RSNA Pneumonia Detection Challenge dataset,\footnote{\url{https://www.kaggle.com/c/rsna-pneumonia-detection-challenge}\label{1}} 2) VinBigData Chest X-ray Abnormalities Detection dataset (VinDr-CXR)\footnote{\url{https://www.kaggle.com/c/vinbigdata-chest-xray-abnormalities-detection}\label{2}} \citep{nguyen2022vindr}, 3) Chest X-ray Anomaly Detection (CXAD) dataset, 4) Brain Tumor MRI dataset,\footnote{\url{https://www.kaggle.com/datasets/masoudnickparvar/brain-tumor-mri-dataset}\label{3}} and 5) Large-scale Attention-based Glaucoma (LAG) dataset \citep{li2019attention}. \\
\textbf{RSNA dataset:} The dataset contains 8851 normal and 6012 lung opacity CXRs. In experiments, we use 3851 normal images as the normal dataset $D_n$, 4000 images with different ARs as the unlabeled dataset $D_u$, and 1000 normal and 1000 lung opacity images as the test dataset $D_t$. \\
\textbf{VinDr-CXR dataset:} The dataset contains 10606 normal and 4394 abnormal CXRs that include 14 categories of anomalies in total. In experiments, we use 4000 normal images as $D_n$, 4000 images as $D_u$, and 1000 normal and 1000 abnormal images as $D_t$. \\
\textbf{CXAD dataset:} The dataset is collected by us for this study, and contains 3299 normal and 1701 abnormal CXRs that include 18 categories of anomalies in total. In experiments, we use 2000 normal images as $D_n$, 2000 images as $D_u$, and 499 normal and 501 abnormal images as $D_t$. \\
\textbf{Brain Tumor MRI dataset:} The dataset contains 2000 MRI slices with no tumors, 1621 with glioma, and 1645 with meningioma. The glioma and meningioma are regarded as anomalies. In experiments, we use 1000 normal images (with no tumors) as $D_n$, 1000 images as $D_u$, and 600 normal and 600 abnormal images (300 with glioma and 300 with meningioma) as $D_t$. \\
\textbf{LAG dataset:} The dataset contains 3143 normal retinal fundus images and 1711 abnormal retinal fundus images with glaucoma. In experiments, we use 1500 normal images as $D_n$, 1500 images as $D_u$, and 811 normal and 811 abnormal images as $D_t$.

We show a summary of the details of the dataset repartitions in Table~\ref{data_repartition}. For the OCC setting, only $D_n$ is used during training. For our proposed training mode, both $D_n$ and $D_u$ are utilized. Except for our CXAD, the reorganized benchmarks and corresponding repartition files have been released for reproducibility. As publicly available benchmarks for anomaly detection in medical images are rare, our released benchmarks will significantly contribute to a fair comparison of studies.

\begin{table*}[!t]
\centering
\caption{Comparison with SOTA methods. For methods that do not use unlabeled images, the two best results are marked in \textbf{bold} and \underline{underlined}. For methods that use unlabeled images, the best results are marked in \underline{\textbf{underlined bold}}. Note that ``IN-Pretr.'' refers to ``ImageNet-Pretrained'', ``Scrat.'' refers to ``trained-from-scratch''  , ``e2e'' refers to end-to-end, and ``*'' refers to incorporating unlabeled data to synthesize anomalies in self-supervised methods.}
\label{sota}
\resizebox{\linewidth}{!}{
\begin{tabular}{c|l|l|cc|cc|cc|cc|cc}
\hline
\multirow{2}{*}{\textbf{Unlabeled data}} &
  \multirow{2}{*}{\textbf{Method}} &
  \multirow{2}{*}{\textbf{Taxonomy}} &
  \multicolumn{2}{c|}{\textbf{RSNA}} &
  \multicolumn{2}{c|}{\textbf{VinDr-CXR}} &
  \multicolumn{2}{c|}{\textbf{CXAD}} &
  \multicolumn{2}{c|}{\textbf{Brain MRI}} &
  \multicolumn{2}{c}{\textbf{LAG}} \\ \cline{4-13} 
 &
   &
   &
  AUC &
  AP &
  AUC &
  AP &
  AUC &
  AP &
  AUC &
  AP &
  AUC &
  AP \\ \hline
\multirow{16}{*}{\ding{55}} &
  AE &
  Rec. &
  66.9 &
  66.1 &
  55.9 &
  60.3 &
  55.6 &
  59.6 &
  79.7 &
  71.9 &
  79.3 &
  76.1 \\
 &
  MemAE \citep{gong2019memorizing}&
  Rec. &
  68.0 &
  67.1 &
  55.8 &
  59.8 &
  56.0 &
  60.0 &
  77.4 &
  70.0 &
  78.5 &
  74.9 \\
 &
  Ganomaly \citep{akcay2018ganomaly}&
  Rec. &
  71.4 &
  69.1 &
  59.6 &
  60.3 &
  62.5 &
  63.0 &
  75.1 &
  69.7 &
  77.7 &
  75.7 \\
 &
  DRAEM \citep{zavrtanik2021draem}&
  Rec.+Self-sup. &
  62.3 &
  61.6 &
  63.0 &
  68.3 &
  54.3 &
  55.6 &
  72.1 &
  64.6 &
  47.2 &
  49.0 \\
 &
  $\text{CutPaste}^\text{IN-Pretr.}$ \citep{li2021cutpaste}&
  Self-sup.+GDE &
  79.4 &
  74.4 &
  70.2 &
  69.8 &
  53.6 &
  57.3 &
  92.0 &
  89.4 &
  69.1 &
  64.6 \\
 &
  $\text{CutPaste}^\text{Scrat.}$ \citep{li2021cutpaste}&
  Self-sup.+GDE &
  75.1 &
  72.6 &
  59.6 &
  58.6 &
  50.3 &
  53.6 &
  92.0 &
  89.9 &
  63.4 &
  59.8 \\
 &
  $\text{CutPaste (e2e)}$ \citep{schluter2022natural}&
  Self-sup. &
  55.0 &
  58.0 &
  54.6 &
  55.5 &
  47.0 &
  48.4 &
  71.0 &
  66.8 &
  53.7 &
  53.9 \\
 &
  FPI \citep{tan2020detecting}&
  Self-sup. &
  47.6 &
  55.7 &
  48.2 &
  49.9 &
  44.8 &
  47.6 &
  83.1 &
  78.9 &
  53.4 &
  55.6 \\
 &
  PII \citep{tan2021detecting}&
  Self-sup. &
  82.9 &
  83.6 &
  65.9 &
  65.8 &
  52.7 &
  53.7 &
  84.3 &
  80.5 &
  61.0 &
  60.7 \\
 &
  NSA \citep{schluter2022natural}&
  Self-sup. &
  82.2 &
  82.6 &
  64.4 &
  65.8 &
  58.5 &
  58.2 &
  84.4 &
  81.1 &
  67.9 &
  67.0 \\
 &
  f-AnoGAN \citep{schlegl2019f}&
  Rec. &
  79.8 &
  75.6 &
  76.3 &
  \textbf{74.8} &
  61.9 &
  \underline{67.3} &
  82.5 &
  74.3 &
  \underline{84.2} &
  77.5 \\
 &
  IGD \citep{chen2022deep}&
  Rec. &
  81.2 &
  78.0 &
  59.2 &
  58.7 &
  55.2 &
  57.6 &
  \textbf{94.3} &
  \underline{90.6} &
  80.7 &
  75.3 \\
 &
  AE-U \citep{mao2020abnormality}&
  Rec. &
  86.7 &
  84.7 &
  73.8 &
  72.8 &
  \underline{66.4} &
  66.9 &
  94.0 &
  89.0 &
  81.3 &
  \underline{78.9} \\ \cline{2-13} 
 &
  Ours (AE), $\mathcal{R}_{intra}$ &
  Ens.+Self-sup. &
  86.3 &
  85.5 &
  \underline{77.2} &
  74.2 &
  63.8 &
  65.4 &
  85.0 &
  77.6 &
  79.5 &
  74.5 \\
 &
  Ours (MemAE), $\mathcal{R}_{intra}$ &
  Ens.+Self-sup. &
  \underline{87.2} &
  \underline{86.1} &
  73.9 &
  72.1 &
  62.4 &
  64.5 &
  82.9 &
  78.6 &
  80.1 &
  77.6 \\
 &
  Ours (AE-U), $\mathcal{R}_{intra}$ &
  Ens.+Self-sup. &
  \textbf{88.3} &
  \textbf{87.6} &
  \textbf{78.2} &
  \underline{74.6} &
  \textbf{69.4} &
  \textbf{69.3} &
  \underline{94.2} &
  \textbf{91.9} &
  \textbf{86.0} &
  \textbf{84.0} \\ \hline
\multirow{6}{*}{\checkmark} &
  CutPaste (e2e)* \citep{schluter2022natural}&
  Self-sup. &
  59.8 &
  61.7 &
  59.2 &
  60.0 &
  48.9 &
  50.7 &
  69.8 &
  64.9 &
  48.9 &
  51.7 \\
 &
  FPI* \citep{tan2020detecting}&
  Self-sup. &
  46.6 &
  53.8 &
  47.4 &
  49.4 &
  45.3 &
  47.6 &
  86.6 &
  83.8 &
  52.9 &
  56.1 \\
 &
  PII* \citep{tan2021detecting}&
  Self-sup. &
  84.3 &
  85.4 &
  66.8 &
  67.2 &
  54.4 &
  54.8 &
  90.0 &
  89.1 &
  63.1 &
  63.1 \\
 &
  NSA* \citep{schluter2022natural}&
  Self-sup. &
  84.2 &
  84.3 &
  64.4 &
  64.8 &
  57.4 &
  57.0 &
  88.8 &
  84.7 &
  68.6 &
  68.0 \\ \cline{2-13} 
 &
  Ours (AE), $\mathcal{R}_{dual}$ &
  Ens.+Self-sup. &
  89.3 &
  89.5 &
  77.4 &
  77.7 &
  65.0 &
  67.2 &
  93.0 &
  87.1 &
  89.0 &
  86.9 \\
 &
  Ours (MemAE), $\mathcal{R}_{dual}$ &
  Ens.+Self-sup. &
  88.5 &
  87.8 &
  75.3 &
  74.1 &
  63.5 &
  64.3 &
  91.4 &
  84.8 &
  88.7 &
  86.5 \\
 &
  Ours (AE-U), $\mathcal{R}_{dual}$ &
  Ens.+Self-sup. &
  \underline{\textbf{91.3}} &
  \underline{\textbf{91.6}} &
  \underline{\textbf{85.9}} &
  \underline{\textbf{84.3}} &
  \underline{\textbf{71.0}} &
  \underline{\textbf{72.7}} &
  \underline{\textbf{97.2}} &
  \underline{\textbf{95.2}} &
  \underline{\textbf{93.1}} &
  \underline{\textbf{92.3}} \\ \hline
\end{tabular}
}
\end{table*}

\subsection{Implementation Details}
The AE in our experiments contains an encoder and a decoder. The encoder contains four convolutional layers with kernel size 4 and stride 2, whose channel sizes are 16-32-64-64. The decoder contains four deconvolutional layers with the same kernel size and stride as the encoder, and the channel sizes are 64-32-16-1. The encoder and decoder are connected by three fully connected layers. All layers except the output layer are followed by batch normalization (BN) and ReLU. For fair comparison, MemAE \citep{gong2019memorizing} and AE-U \citep{mao2020abnormality} are modified in our experiments based on this AE. All the input images are resized to $64 \times 64$, $K$ is set to 3, and all the reconstruction models are trained for 250 epochs using the Adam optimizer with a learning rate of 5e-4. 

The proposed ASR-Net consists of three cascaded convolutional layers, connected by BN and ReLU. It is trained for 100 epochs with a learning rate of 1e-4 and a weight decay of 1e-4 to ensure convergence.

All experiments are implemented using PyTorch. The performance is assessed with the area under the ROC curve (AUC) and average precision (AP).

\subsection{Comparison with State-of-the-art Methods}
\label{sec:exp_c}

In Table~\ref{sota}, we compare our proposed method with a wide range of state-of-the-art (SOTA) methods, including MemAE \citep{gong2019memorizing}, Ganomaly \citep{akcay2018ganomaly}, DRAEM \citep{zavrtanik2021draem}, CutPaste (including ImageNet-pretrained and trained-from-scratch versions) \citep{li2021cutpaste}, CutPaste (e2e) \citep{schluter2022natural}, FPI \citep{tan2020detecting}, PII \citep{tan2021detecting}, NSA \citep{schluter2022natural}, f-AnoGAN \citep{schlegl2019f}, IGD \citep{chen2022deep} and AE-U \citep{mao2020abnormality}. Note that the official code of CutPaste \citep{li2021cutpaste} has not been released. Thus, we use a public implementation from \url{https://github.com/Runinho/pytorch-cutpaste}. For fair comparison among standard self-supervised methods, we use the unified implementation provided by NSA \citep{schluter2022natural} for CutPaste (e2e), FPI, and PII. All other methods used in the experiments are implemented using their official codes.

\subsubsection{Performance under the OCC setting}
We compare our DDAD-$\mathcal{R}_{intra}$ with others under the same OCC setting for fairness; i.e., only the normal dataset $D_n$ is used during training without the use of unlabeled images. Under the OCC setting, the two best results are marked in \textbf{bold} and \underline{underlined} in Table~\ref{sota}. The results show that our DDAD built on AE-U using $\mathcal{R}_{intra}$ as the AS achieves SOTA results on almost all the five benchmarks comprising three different medical image modalities (CXR, brain MRI and retinal fundus image), demonstrating the effectiveness and generality of our proposed method. Our method also outperforms other SOTA self-supervised methods (e.g., NSA \citep{schluter2022natural}). However, FPI \citep{tan2020detecting}, with the same synthesis approach as ours, performs poorly on the five datasets. The reason is that FPI \citep{tan2020detecting} and other similar self-supervised methods overfit the synthetic anomalies. In contrast, our ASR-Net never sees the synthetic anomalies, and instead takes the anomaly score maps as input to learn the refinement, avoiding the overfitting problem. Specifically, standard self-supervised methods achieve satisfactory performance on the Brain Tumor MRI dataset, where the anomalies (tumors) present a notable intensity discrepancy from the normal regions, similar to the synthetic abnormal patterns. However, the predominant manifestation of abnormal (glaucoma) images in the LAG dataset \citep{li2019attention} is alterations in the optic disk appearance and vasculature, which differ significantly from the synthetic abnormal patterns. As a result, standard self-supervised methods fail to detect these anomalies, while in our proposed method, anomaly cues are effectively captured by DDAD and refined by our ASR-Net, resulting in accurate predicted abnormal regions.

Another surprising observation is that MemAE \citep{gong2019memorizing} often performs worse than AE. The reason could be that the difference between normal and abnormal medical images is significantly smaller than that between natural images in the original paper of MemAE \citep{gong2019memorizing}. In medical domains, abnormal images always contain only subtle lesions to differentiate them from normal images, and their features can be easily obtained using the combination of some normal features, as they are similar overall.

\subsubsection{Performance under the OC-SSL setting}
We evaluate the proposed method in the situation that the unlabeled image dataset $D_u$ is utilized, i.e., use $\mathcal{R}_{dual}$ as the AS. Referencing the ARs of several public medical image datasets (e.g., 71\% in RSNA, 46\% in ChestX-ray8 \citep{wang2017chestx} and 62\% in Brain Tumor MRI), we generally assume an AR of 60\% for $D_u$ in the experiments. For fair comparison, we incorporate the unlabeled dataset for other self-supervised methods to synthesize more diverse anomalies in training. Under this setting, the best results are marked in \underline{\textbf{underlined bold}} in Table~\ref{sota}. While our DDAD (AE-U) using $\mathcal{R}_{intra}$ achieves SOTA results, our $\mathcal{R}_{dual}$ further improves the performance with the help of unlabeled images, outperforming the previous methods by a larger margin. For other self-supervised methods, including CutPaste (e2e) \citep{schluter2022natural}, FPI \citep{tan2020detecting}, PII \citep{tan2021detecting} and NSA \citep{schluter2022natural}, some performance improvement is obtained from the unlabeled data, but it is overall limited. These results indicate that our proposed method is able to more effectively capture useful information from unlabeled images for anomaly detection.

\subsection{Ablation Study}
\label{sec:exp_ablation}

\subsubsection{DDAD with different ARs}

\begin{figure}[!t]
    \centering
    \includegraphics[width=\linewidth]{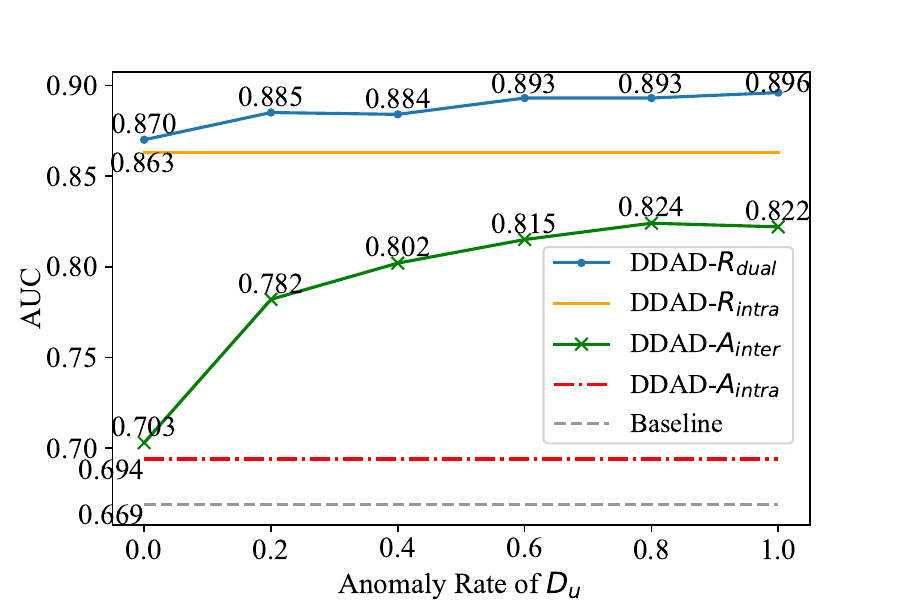}
    \caption{Performance of DDAD and the reconstruction baseline on the RSNA dataset with a varying AR of $D_u$ using AE as the backbone.}
    \label{Ablation_AR}
\end{figure}

In clinical practice, the AR of unlabeled dataset $D_u$ is unknown. In order to simulate various real scenarios, we evaluate the proposed DDAD on the RSNA dataset with the AR of $D_u$ varying from 0 to 100\%. We use the reconstruction method as the baseline for comparison. For fair comparison, all these methods use AE as the backbone. The results of proposed DDAD method using $\mathcal{R}_{dual}$, $\mathcal{R}_{intra}$, $\mathcal{A}_{inter}$ and $\mathcal{A}_{intra}$, and the results of reconstruction baseline are shown in Fig.~\ref{Ablation_AR}. They clearly demonstrate the effectiveness of our proposed anomaly scores and ASR-Net.

\begin{table*}[!b]
\centering
\caption{Performance of different methods using three backbones on five datasets. The best two results for each backbone are marked in \underline{\textbf{underlined bold}} and \textbf{bold}.}
\label{Ablation_backbone}
\resizebox{\linewidth}{!}{
\begin{tabular}{c|c|ccccccccccccccc}
\hline
\multirow{3}{*}{\textbf{Method}} &
  \multirow{3}{*}{\textbf{AS}} &
  \multicolumn{15}{c}{\textbf{AUC}} \\ \cline{3-17} 
 &
   &
  \multicolumn{3}{c|}{RSNA} &
  \multicolumn{3}{c|}{VinDr-CXR} &
  \multicolumn{3}{c|}{CXAD} &
  \multicolumn{3}{c|}{Brain MRI} &
  \multicolumn{3}{c}{LAG} \\ \cline{3-17} 
 &
   &
  AE &
  MemAE &
  \multicolumn{1}{c|}{AE-U} &
  AE &
  MemAE &
  \multicolumn{1}{c|}{AE-U} &
  AE &
  MemAE &
  \multicolumn{1}{c|}{AE-U} &
  AE &
  MemAE &
  \multicolumn{1}{c|}{AE-U} &
  AE &
  MemAE &
  AE-U \\ \hline
Rec. &
  \multirow{2}{*}{$\mathcal{A}_{rec}$} &
  66.9 &
  68.0 &
  \multicolumn{1}{c|}{86.7} &
  55.9 &
  55.8 &
  \multicolumn{1}{c|}{73.8} &
  55.6 &
  56.0 &
  \multicolumn{1}{c|}{66.4} &
  79.7 &
  77.4 &
  \multicolumn{1}{c|}{94.0} &
  79.3 &
  78.5 &
  81.3 \\
Rec. (ensemble) &
   &
  66.9 &
  67.0 &
  \multicolumn{1}{c|}{86.6} &
  55.5 &
  55.3 &
  \multicolumn{1}{c|}{73.1} &
  55.0 &
  55.2 &
  \multicolumn{1}{c|}{65.9} &
  81.3 &
  79.2 &
  \multicolumn{1}{c|}{93.3} &
  78.8 &
  79.2 &
  82.1 \\ \hline
\multirow{4}{*}{DDAD} &
  $\mathcal{A}_{intra}$ &
  69.4 &
  72.9 &
  \multicolumn{1}{c|}{87.3} &
  60.1 &
  59.5 &
  \multicolumn{1}{c|}{74.3} &
  59.8 &
  59.4 &
  \multicolumn{1}{c|}{69.2} &
  55.9 &
  52.6 &
  \multicolumn{1}{c|}{84.5} &
  72.1 &
  71.3 &
  75.3 \\
 &
  $\mathcal{A}_{inter}$ &
  81.5 &
  78.8 &
  \multicolumn{1}{c|}{\textbf{91.0}} &
  71.0 &
  69.0 &
  \multicolumn{1}{c|}{\underline{\textbf{85.9}}} &
  62.1 &
  59.9 &
  \multicolumn{1}{c|}{\underline{\textbf{71.4}}} &
  84.4 &
  \textbf{83.2} &
  \multicolumn{1}{c|}{\textbf{97.1}} &
  \textbf{87.2} &
  \textbf{88.5} &
  80.6 \\ \cline{2-17} 
 &
  $\mathcal{R}_{intra}$ &
  \textbf{86.3} &
  \textbf{87.2} &
  \multicolumn{1}{c|}{88.3} &
  \textbf{77.2} &
  \textbf{73.9} &
  \multicolumn{1}{c|}{78.2} &
  \textbf{63.8} &
  \textbf{62.4} &
  \multicolumn{1}{c|}{69.4} &
  \textbf{85.0} &
  82.9 &
  \multicolumn{1}{c|}{94.2} &
  79.5 &
  80.1 &
  \textbf{86.0} \\
 &
  $\mathcal{R}_{dual}$ &
  \underline{\textbf{89.3}} &
  \underline{\textbf{88.5}} &
  \multicolumn{1}{c|}{\underline{\textbf{91.3}}} &
  \underline{\textbf{77.4}} &
  \underline{\textbf{75.3}} &
  \multicolumn{1}{c|}{\underline{\textbf{85.9}}} &
  \underline{\textbf{65.0}} &
  \underline{\textbf{64.5}} &
  \multicolumn{1}{c|}{\textbf{71.0}} &
  \underline{\textbf{93.0}} &
  \underline{\textbf{91.4}} &
  \multicolumn{1}{c|}{\underline{\textbf{97.2}}} &
  \underline{\textbf{89.0}} &
  \underline{\textbf{88.7}} &
  \underline{\textbf{93.1}} \\ \hline
\end{tabular}
}
\end{table*}

Firstly, DDAD using the original $\mathcal{A}_{intra}$ and $\mathcal{A}_{inter}$ achieves consistent and significant improvement compared with the reconstruction baseline, suggesting that the two proposed  ASs are more discriminative than the previous reconstruction error. Moreover, our $\mathcal{A}_{inter}$ is better than $\mathcal{A}_{intra}$, while it performs better with an increasing AR of $D_u$, consistent with our hypothesis in Section \ref{AS} that a higher AR of $D_u$ will result in a more competitive $\mathcal{A}_{inter}$. Because $\mathcal{A}_{intra}$ is computed inside the NDM, it is irrelevant to the AR. It is worth noting that even in the extreme situation that AR is 0, our DDAD-$\mathcal{A}_{inter}$ can still achieve better performance than baseline. That is to say, we can apply the DDAD in any situations and get improvement, regardless of the AR. Intuitively, when the AR is 0, dataset $D_n \cup D_u$ only contains normal images, and thus the UDM degenerates to be the same as the NDM. However, in this situation the UDM is trained on a larger normal dataset than baseline, which leads to more robust models and supports the consistent improvement. Meanwhile, even if the AR is low (e.g., 20\%), the DDAD can achieve a significant improvement (7.9\% AUC higher than when the AR is 0). That means the proposed DDAD can improve the performance considerably in clinical practice as there are always some abnormal cases.

Secondly, refined by the proposed ASR-Net, our $\mathcal{R}_{dual}$ and $\mathcal{R}_{intra}$ have a further significant gain compared with the original $\mathcal{A}_{inter}$ and $\mathcal{A}_{intra}$. Specifically, when using only normal images, our ASR-Net $F(\cdot)$ refines $\mathcal{A}_{intra}$ and derives $\mathcal{R}_{intra}$, which improves the AUC of $\mathcal{A}_{intra}$ by a large margin of 16.9\% (from 69.4\% to 86.3\%). Incorporating the unlabeled images, we can derive $\mathcal{A}_{inter}$ as a complement to $\mathcal{A}_{intra}$. The two ASs are refined and fused by $F(\cdot)$, deriving $\mathcal{R}_{dual}$, which achieves an AUC of 87.0\%--89.6\% with the AR of $D_u$ varying from 0 to 100\%, outperforming all the aforementioned methods. More importantly, while our $\mathcal{R}_{dual}$ utilizes unlabeled images and achieves advanced performance, it is insensitive to the AR of $D_u$. Even if the AR is 0, it can achieve an AUC of 87.0\%, which outperforms $\mathcal{A}_{inter}$ in any situations. Therefore, we can conclude that with the help of ASR-Net, the DDAD is more robust and it can handle various complex situations in clinical practice well.

\subsubsection{DDAD with different backbones}

Our proposed DDAD method can use any variants of AE as the backbone. To further prove its superiority, DDAD built on different backbones is compared with the corresponding reconstruction baselines (Rec.) in Table~\ref{Ablation_backbone}. The two best results for each backbone are marked in \underline{\textbf{underlined bold}} and \textbf{bold}. Consistent with Section~\ref{sec:exp_c}, we also assume an AR of 60\% for $D_u$ in experiments. The results show that DDAD based on AE, MemAE \citep{gong2019memorizing} and AE-U \citep{mao2020abnormality} can all outperform the corresponding baselines on the five datasets by a large margin.

Specifically, all of our original $\mathcal{A}_{intra}$ and $\mathcal{A}_{inter}$, and the refined $\mathcal{R}_{intra}$ and $\mathcal{R}_{dual}$ perform competitively on the three CXR datasets (RSNA, VinDr-CXR and CXAD datasets). In terms of AUC, DDAD-$\mathcal{A}_{intra}$ improves on the baselines AE, MemAE and AE-U by 2.5\%, 4.9\% and 0.6\% on the RSNA dataset, 4.2\%, 3.7\% and 0.5\% on the VinDr-CXR dataset, 4.2\%, 3.4\% and 2.8\% on the CXAD dataset. DDAD-$\mathcal{A}_{inter}$ improves on the same baselines by 14.6\%, 10.8\% and 4.3\% on the RSNA dataset, 15.1\%, 13.2\% and 12.1\% on the VinDr-CXR dataset, 6.5\%, 3.9\% and 5.0\% on the CXAD dataset. With the help of our ASR-Net, DDAD-$\mathcal{R}_{intra}$ improves the baselines AE, MemAE and AE-U by 19.4\%, 19.2\% and 1.6\% on the RSNA dataset, 21.3\%, 18.1\% and 4.4\% on the VinDr-CXR dataset, 8.2\%, 6.4\% and 3.0\% on the CXAD dataset, while for DDAD-$\mathcal{R}_{dual}$, the improvement is 22.4\%, 20.5\% and 4.6\% on the RSNA dataset, 21.5\%, 19.5\% and 12.1\% on the VinDr-CXR dataset, 9.4\%, 7.5\% and 4.6\% on the CXAD dataset.

As for the Brain MRI and LAG dataset, the proposed original $\mathcal{A}_{intra}$ performs worse than the corresponding reconstruction baseline. However, with the aid of our ASR-Net, $\mathcal{R}_{intra}$ significantly improves the performance of $\mathcal{A}_{intra}$ and outperforms the corresponding baseline by a large margin. The reason could be that, although the original $\mathcal{A}_{intra}$ contains noises and works unsatisfactorily, it does encode useful information for anomaly detection, which is successfully extracted by our ASR-Net, deriving the $\mathcal{R}_{intra}$. Finally, consistent with the results on the three CXR datasets, our refined $\mathcal{R}_{intra}$ and $\mathcal{R}_{dual}$ outperform the original $\mathcal{A}_{intra}$ and $\mathcal{A}_{inter}$ on the Brain Tumor and LAG datasets, while showing their superiority to reconstruction baselines. 

We also test the ensemble of $K$ reconstruction models using $\mathcal{A}_{rec}$, shown as ``Rec. (ensemble)'' in Table~\ref{Ablation_backbone}, demonstrating that a simple ensemble has no significant improvement. The reason why some ensembles result in slightly worse performance could be that the average reconstruction of ensemble networks may generalize better than the single network on some abnormal regions, causing reconstruction errors in these regions to be indistinguishable from those of normal regions.

\subsubsection{Performance on seen and unseen pathologies}

\begin{table*}[hb]
\centering
\caption{Performance of DDAD on seen and unseen pathologies. Setting A indicates the testing set contains only pathologies in $\mathcal{P}_A$, which could appear in $D_u$. Setting B indicates the testing set contains only pathologies in $\mathcal{P}_B$, which are unseen in $D_u$.}
\label{exp_unseen}
\begin{tabular}{l|l|ll|ll}
\hline
\multirow{2}{*}{Method}                     & \multirow{2}{*}{Unlabeled dataset $D_u$}                 & \multicolumn{2}{c|}{Setting A}          & \multicolumn{2}{c}{Setting B}             \\ \cline{3-6} 
                                            &                                                   & AUC (\%)              & AP (\%)               & AUC (\%)              & AP (\%)               \\ \hline
Reconstruction                              & 0                                                 & 49.7                  & 55.6                  & 63.7                  & 70.1                  \\ \hline
\multirow{2}{*}{DDAD-$\mathcal{A}_{inter}$} & 4000 (4000 normal + 0 abnormal images)                          & 54.0                  & 60.0                  & 66.0                  & 71.1                  \\
                                            & 4000 (2412 normal + 1588 abnormal images in $\mathcal{P}_A$)  & $64.2^{+10.2}$  & $70.8^{+10.8}$  & $70.0^{+4.0}$   & $75.8^{+4.7}$   \\ \hline
\end{tabular}
\end{table*}

\begin{figure}[!t]
\centering
\includegraphics[width=\linewidth]{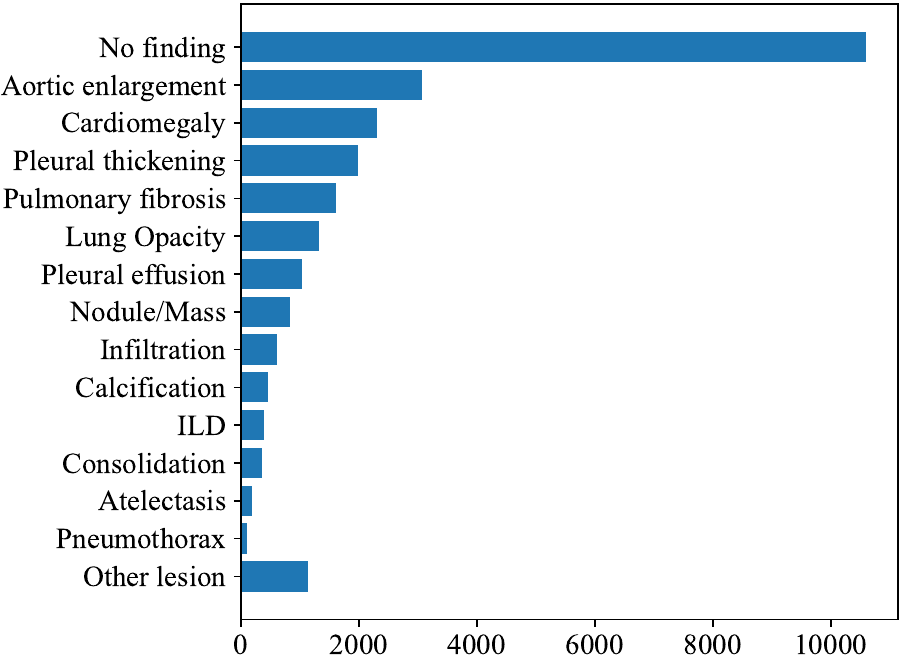}
\caption{Class distribution of the VinDr-CXR dataset. Each abnormal image could contain multiple categories of diseases.} \label{Vin-Dr}
\end{figure}

In clinical practice, the recognition of rare diseases is an important but very intractable task, where even unlabeled samples containing certain rare diseases are infeasible to acquire. Therefore, exploring our method's performance under the situation that the unlabeled dataset $D_u$ contains multiple diseases while the testing set contains different types of unseen diseases is meaningful. To simulate this situation and evaluate our method on seen and unseen pathologies, we utilize the VinDr-CXR dataset, which contains various types of pathologies as shown in Fig.~\ref{Vin-Dr}. We define a set of several pathologies, $\mathcal{P}_A$ = \{aortic enlargement, cardiomegaly, lung opacity, pleural thickening, pulmonary fibrosis\}, which contains the five most common pathologies in the dataset, as the seen pathologies to build the unlabeled dataset $D_u$ for training. For the unseen pathologies, we use the set of remaining less frequent pathologies, $\mathcal{P}_B$=\{atelectasis, calcification, consolidation, ILD, infiltration, nodule/mass, pleural effusion, pneumothorax, other lesion\}.

We incorporate 1588 abnormal images containing a subset of diseases in $\mathcal{P}_A$ and 2412 normal images as $D_u$. For testing, we utilize 100 normal images, along with 100 abnormal images containing a subset of diseases in $\mathcal{P}_A$ to evaluate the improvement on seen pathologies (Setting A), or 101 abnormal images containing a subset of diseases in $\mathcal{P}_B$ to evaluate the improvement on unseen pathologies (Setting B). As the control group, $\mathcal{A}_{inter}$ trained on the unlabeled dataset $D_u$ that contains only normal images is also evaluated. 

The results are shown in Table~\ref{exp_unseen}. It indicates that when incorporating abnormal images into the unlabeled set $D_u$, DDAD-$\mathcal{A}_{inter}$ has an improvement of 10.2\% AUC and 10.8\% AP on the seen pathologies (Setting A), while an improvement of 4.0\% AUC and 4.7\% AP is also achieved on even the unseen pathologies (Setting B). This reveals the tremendous potential of DDAD for improving the recognition of rare diseases, even if samples containing such diseases are unavailable in the unlabeled dataset.

\subsection{Qualitative Analysis}

\begin{figure*}[!t]
\centering
\subfigure[Reconstruction]{
\begin{minipage}[t]{0.185\linewidth}
    \centering
	\includegraphics[width=\linewidth]{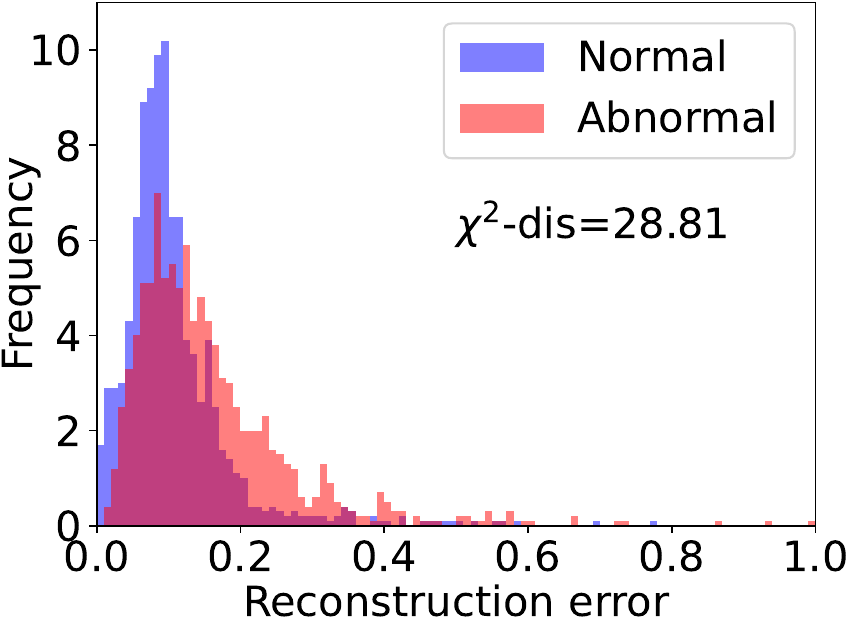}
\end{minipage}%
}
\subfigure[DDAD-$\mathcal{A}_{intra}$]{
\begin{minipage}[t]{0.185\linewidth}
	\centering
	\includegraphics[width=\linewidth]{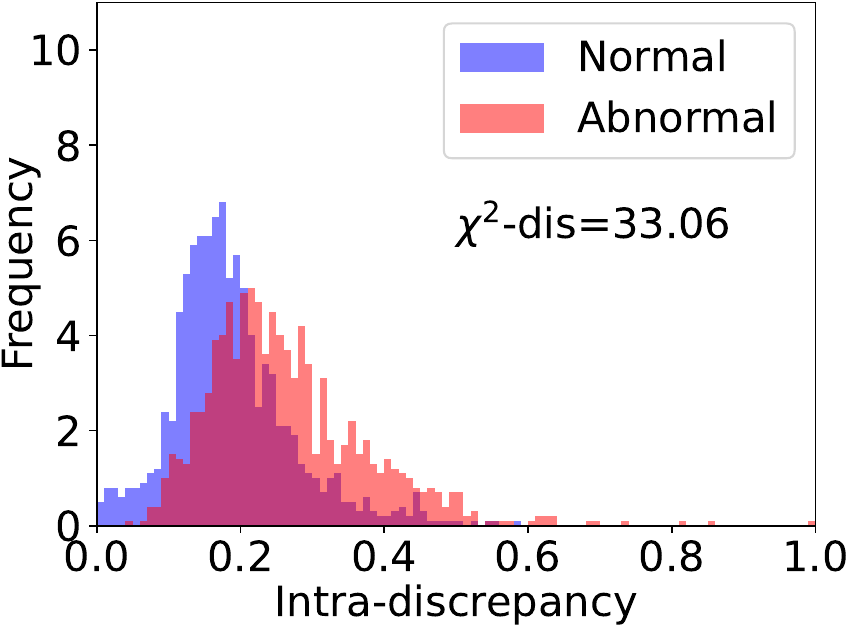}
\end{minipage}
}
\subfigure[DDAD-$\mathcal{A}_{inter}$]{
\begin{minipage}[t]{0.185\linewidth}
	\centering
	\includegraphics[width=\linewidth]{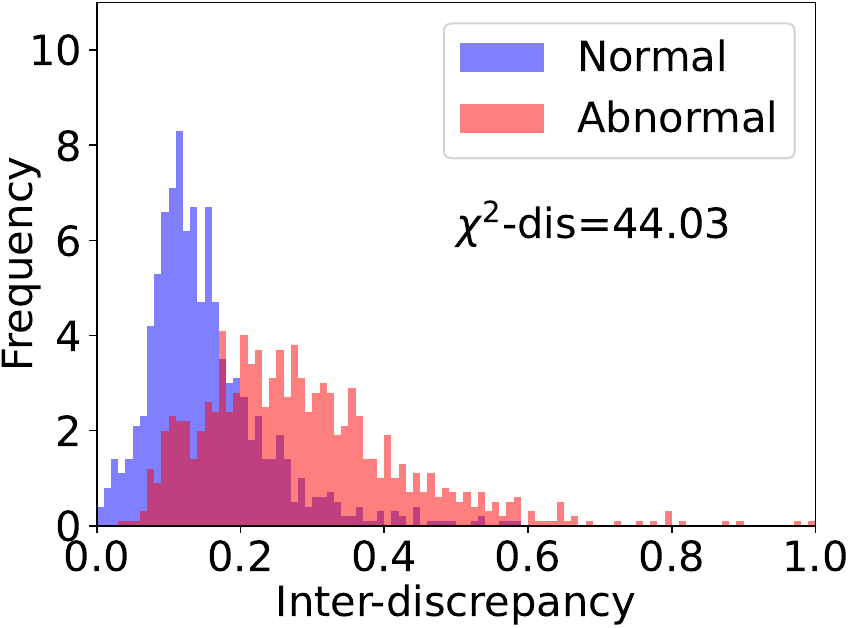}
\end{minipage}
}
\subfigure[DDAD-$\mathcal{R}_{intra}$]{
\begin{minipage}[t]{0.185\linewidth}
	\centering
	\includegraphics[width=\linewidth]{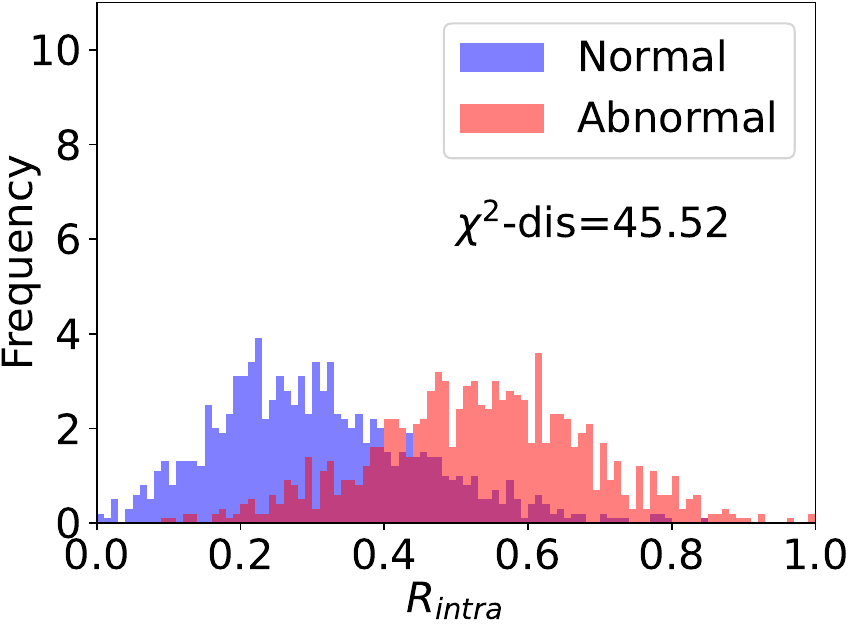}
\end{minipage}
}
\subfigure[DDAD-$\mathcal{R}_{dual}$]{
\begin{minipage}[t]{0.185\linewidth}
	\centering
	\includegraphics[width=\linewidth]{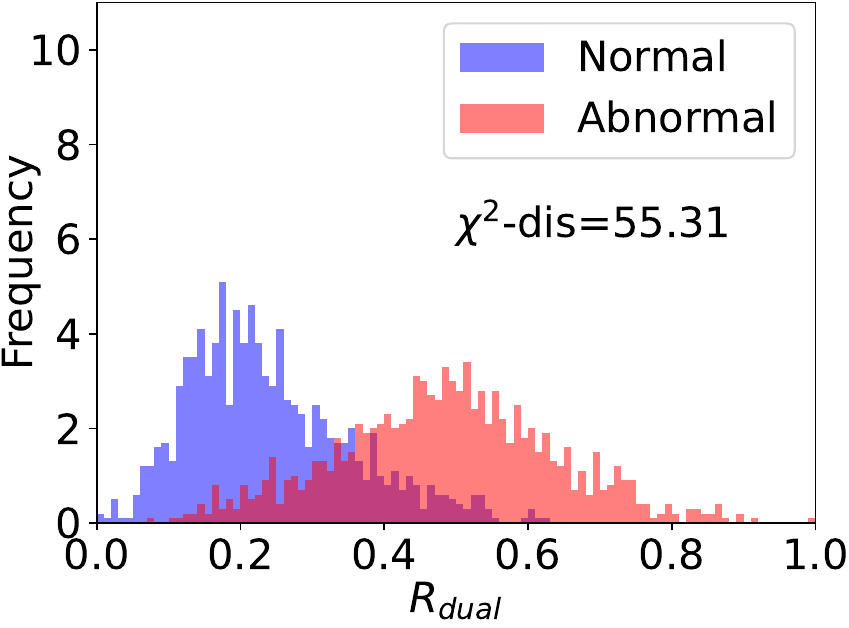}
\end{minipage}
}
\caption{Histograms of anomaly score for normal (blue) and abnormal (red) images in the test set of RSNA. The backbone is AE. Scores are normalized to [0,1]. The $\chi^2$-distance measures the difference between the histograms of normal and abnormal images.}
\label{histogram}
\end{figure*}

To further illustrate the superiority of the proposed method, we conduct qualitative analysis on the RSNA dataset in this section using AS histograms and heat maps.

\subsubsection{AS histograms} To show the discriminative capability of different methods, we visualize the histograms of their AS for normal and abnormal images in Fig.~\ref{histogram} using AE as the backbone. The overlaps of normal and abnormal histograms indicate samples with the same AS but different categories; thus, they are indistinguishable. The $\chi^2$-distance shown in the figure measures the difference between the histograms of normal and abnormal images. Therefore, a larger difference between the ASs of normal and abnormal images will result in fewer overlaps and a larger $\chi^2$-distance, indicating stronger discriminative capability. Based on these analyses and observations, we can draw the conclusion that the proposed DDAD is superior to previous reconstruction methods and our ASR-Net is effective. The performance of different methods (ASs) can be ranked from better to worse as: $\mathcal{R}_{dual}$ and $\mathcal{R}_{intra}$ \textgreater $\mathcal{A}_{inter}$ and $\mathcal{A}_{intra}$ \textgreater $\mathcal{A}_{rec}$, which is consistent with our experimental results.

\begin{figure*}[!t]
\centering
\includegraphics[width=\linewidth]{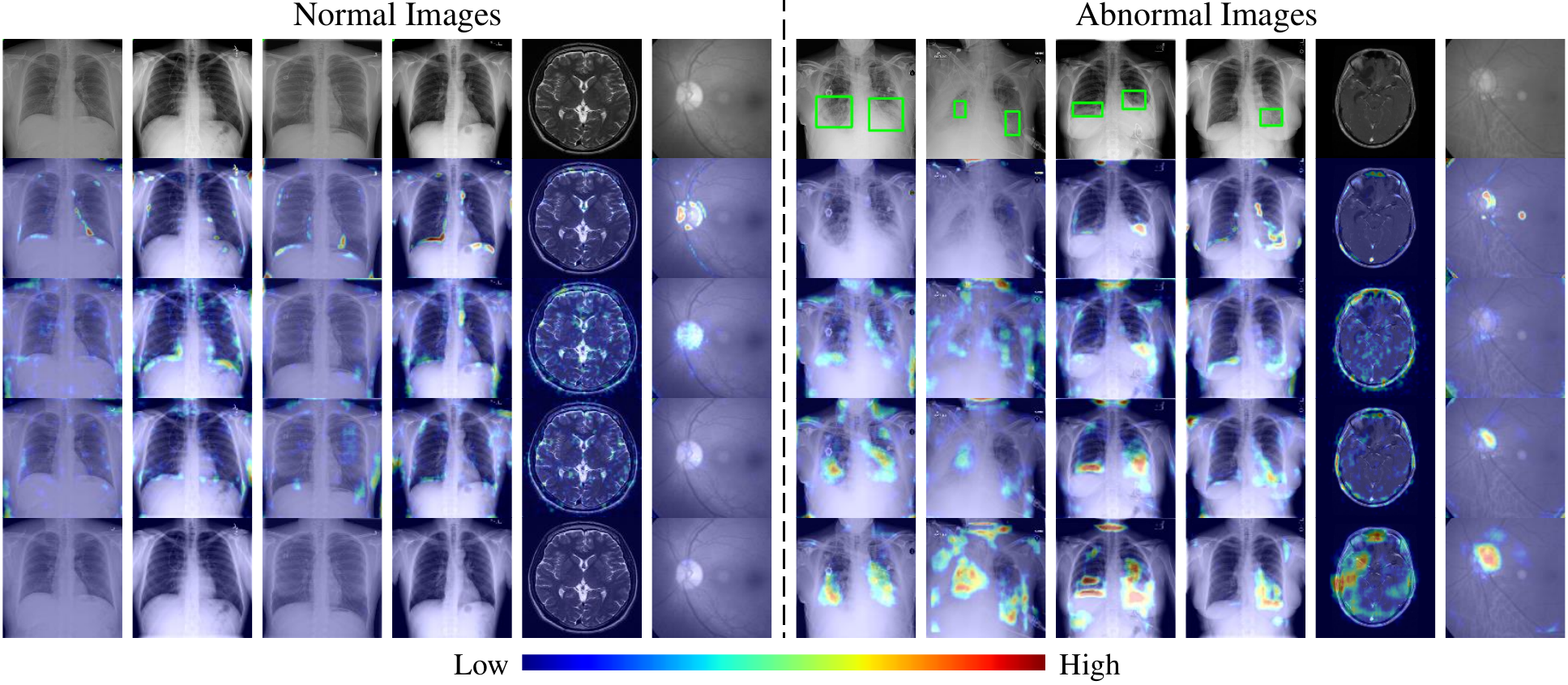}
\caption{Visualization of heat maps on medical datasets. From top to bottom: Original images, heat maps of $\mathcal{A}_{rec}$, heat maps of $\mathcal{A}_{intra}$, heat maps of $\mathcal{A}_{inter}$, heat maps of $\mathcal{R}_{dual}$. The green bounding boxes indicate abnormal regions.}
\label{visualization}
\end{figure*}

\subsubsection{Heat maps} We visualize heat maps of $\mathcal{A}_{rec}$, $\mathcal{A}_{intra}$, $\mathcal{A}_{inter}$, and $\mathcal{R}_{dual}$ on CXRs, brain MRIs, and retinal fundus images for comparison. In Fig.~\ref{visualization}, the previous reconstruction method (in the second row) cannot identify subtle lesions well, while it always has false positives around the normal regions' boundaries. The two proposed discrepancy scores (in the third and fourth row), especially $\mathcal{A}_{inter}$ (in the fourth row), show better discriminative capability to recognize most abnormal regions. With the ASR-Net, our $\mathcal{R}_{dual}$ further remove the false positives of $\mathcal{A}_{intra}$ and $\mathcal{A}_{inter}$ in normal images, while its responses on abnormal regions are enhanced. It can thus perform as a rough localization result for radiologists to reference.


\section{Discussion} \label{sec:Discussion}

\subsection{Impact of the ensemble size $K$}
To analyze the impact of ensemble size $K$ in DDAD, a series of experiments are conducted on the RSNA dataset. As shown in Table~\ref{tab_k}, results suggest that $\mathcal{A}_{intra}$ is insensitive to $K$, while the performance of $\mathcal{A}_{inter}$ first increases and then gradually becomes stable as $K$ increases. Considering that a small $K$ is sufficient to demonstrate the effectiveness of our method, and achieving better performance via a larger ensemble is not our main purpose, we simply choose $K=3$ as a compromise between computational cost and performance.

\begin{table*}[!t]
\centering
\caption{Impact of the ensemble size $K$. The performance is shown in the format AUC(\%)/AP(\%).} \label{tab_k}
\begin{tabular}{cc|cccccc}
\hline
Backbone            & AS                    & $K=2$     & $K=3$     & $K=5$     & $K=7$     & $K=11$    & $K=15$      \\ \hline
\multirow{2}{*}{AE} & $\mathcal{A}_{intra}$ & 69.5/69.3 & 69.4/68.5 & 69.5/68.9 & 69.7/69.2 & 69.0/68.4 & 69.1/68.5 \\
                    & $\mathcal{A}_{inter}$ & 79.6/79.3 & 81.5/81.0 & 84.2/83.4 & 84.8/83.9 & 85.4/84.6 & 86.0/85.1 \\ \hline
\end{tabular}
\end{table*}

\subsection{Uncertainty estimates}
Other than Deep Ensemble, well-known methods for uncertainty estimates include Monte Carlo (MC) Dropout \citep{gal2016dropout}, which is also simple and widely-used. MC Dropout has less computational cost compared with Deep Ensemble but the standard training and testing progress needs to be modified by randomly deactivating some neurons, while Deep Ensemble has better scalability and better performance, requiring few/no modifications to the standard learning progress of the network \citep{lakshminarayanan2017simple}. 

The performances of using Deep Ensemble or MC Dropout in DDAD are shown in Table~\ref{tab_ue}. The results indicate that Deep Ensemble can outperform MC Dropout consistently on both AUC and AP. More importantly, benefiting from the good scalability of Deep Ensemble, the powerful AE-U can be easily applied as our backbone. In contrast, it does not work well when MC Dropout is used. The reason could be that random dropout disturbs the prediction of the automatically learned pixel-level uncertainty map in AE-U, and thereby leads to serious performance deterioration.

\begin{table}[!t]
\centering
\caption{Comparison of Deep Ensemble and MC Dropout for uncertainty estimates in DDAD. Here Deep Ensemble uses an ensemble of three networks, while the MC Dropout executes the forward pass 256 times with random dropout for Monte Carlo estimates.} \label{tab_ue}
\resizebox{\linewidth}{!}{
\begin{tabular}{llccc}
\hline
Backbone              & AS                                     & Uncertainty Estimates & AUC (\%) & \multicolumn{1}{l}{AP (\%)} \\ \hline
\multirow{4}{*}{AE}   & \multirow{2}{*}{$\mathcal{A}_{intra}$} & Deep Ensemble         & 69.4     & 68.5                        \\
                      &                                        & MC Dropout            & 69.5     & 67.8                        \\ \cline{2-5} 
                      & \multirow{2}{*}{$\mathcal{A}_{inter}$} & Deep Ensemble         & 81.5     & 81.0                        \\
                      &                                        & MC Dropout            & 78.5     & 77.0                        \\ \hline
\multirow{4}{*}{AE-U} & \multirow{2}{*}{$\mathcal{A}_{intra}$} & Deep Ensemble         & 87.3     & 86.3                        \\
                      &                                        & MC Dropout            & 63.1     & 63.4                        \\ \cline{2-5} 
                      & \multirow{2}{*}{$\mathcal{A}_{inter}$} & Deep Ensemble         & 91.0     & 91.3                        \\
                      &                                        & MC Dropout            & 79.8     & 81.2                        \\ \hline
\end{tabular}
}
\end{table}


\subsection{Self-supervised learning for anomaly detection}

Self-supervised learning-based methods have become very popular for anomaly detection \citep{li2021cutpaste,zavrtanik2021draem,tan2020detecting,tan2021detecting,schluter2022natural}, and some achieve extremely high performance in industrial or medical applications. However, in the industrial domain, most of the methods are evaluated only on the MVTec AD dataset \citep{bergmann2019mvtec}, which could be insufficient as it is quite possible to synthesize defects in specific patterns that are very helpful for the recognition of anomalies in a specific test set, but not useful for other anomalies. In the medical domain, due to the lack of publicly available benchmarks, previous methods are evaluated on different datasets, hindering comprehensive and fair comparison. These hinder the reasonable analysis of self-supervised methods and restrict the development of anomaly detection. To analyze these methods better and reveal future trends, we compare various methods comprehensively and fairly on five medical datasets, as shown in Table~\ref{sota}.

Surprisingly, our comparison reveals that, although self-supervised methods can perform well on specific datasets, they always fail on other datasets. For example, DRAEM \citep{zavrtanik2021draem} achieves an image-level AUC of 98.0\% on the MVTec AD dataset. However, it performs even worse than the vanilla AE on four of the five medical datasets. NSA \citep{schluter2022natural}, the SOTA self-supervised method, also performs worse than the vanilla AE on the LAG dataset. Meanwhile, several reconstruction-based methods (e.g., AE-U \citep{mao2020abnormality} and f-AnoGAN \citep{schlegl2019f}) show more competitive results than the self-supervised methods on all five datasets. The reason is that most self-supervised methods essentially try to synthesize anomalies inherently similar to the real anomalies in specific datasets. They overfit the synthetic anomalies and cannot recognize real anomalies that are inherently different from their synthetic ones.
Although NSA \citep{schluter2022natural} is designed with some strategies to synthesize more natural and relevant anomalies and outperforms other self-supervised methods, it does not solve this problem and still performs poorly on the LAG dataset. In contrast, reconstruction-based methods recognize deviations from the normal pattern as anomalies, where different anomalies are treated equivalently, thus performing robustly on different datasets. Therefore, in the situations where abnormal patterns are unknown, reconstruction-based methods may be a better choice compared with self-supervised ones.

Although the standard self-supervised methods suffer from overfitting, the results in Section~\ref{Experiments} reveal that using self-supervised learning for refinement or representation learning can achieve better performance. Table~\ref{sota} and \ref{Ablation_backbone} show that our ASR-Net for self-supervised refinement significantly improves the performance on the five benchmarks based on the three backbones. However, FPI \citep{tan2020detecting}, using the same synthesis approach as ours, performs worse than ours on all five datasets. This phenomenon is highly related to what networks learn through self-supervised learning. The standard self-supervised methods directly learn to detect synthetic abnormal patterns, and thus easily overfit. In contrast, our ASR-Net learns the mapping function from the original AS to the final accurate abnormal regions, which are unrelated to the abnormal patterns, and thus generalizes well to anomalies in various scenarios. 

Moreover, $\text{CutPaste}^\text{Scrat.}$ \citep{li2021cutpaste}, which builds a Gaussian density estimator (GDE) \citep{rippel2021modeling} on learned representations, outperforms CutPaste (e2e) \citep{schluter2022natural} by a large margin on all five datasets. This reveals that although the synthetic abnormal patterns are not a perfect simulation of real anomalies, training the network to classify them is able to learn representations that can distinguish between normality and real abnormality. Therefore, using self-supervised representation is more promising than using the network trained via self-supervised learning to directly detect anomalies.

In summary, compared with standard self-supervised methods that focus on training the network to directly detect anomalies, designing self-supervised tasks like refinement and representation learning that are insensitive to abnormal patterns is more generalizable, promising and competitive in complex real scenarios.

\subsection{Limitations}
Currently, our ASR-Net does have limitations. In the experiments, it shows only a small improvement when the original dual-distribution discrepancy refined by the uncertainty from AE-U has already achieved a high performance (i.e., DDAD (AE-U) in Table~\ref{Ablation_backbone}). The reason could be that our refinement strategy is conducted on the discrepancy maps of ensembles of reconstruction networks, causing the upper bound of performance to be limited by the distribution-modeling capability of these reconstruction networks. Therefore, some subtle abnormal regions that are reconstructed consistently by different networks in the ensemble are unable to be recognized, regardless of the subsequent refinement. In future work, we intend to explore a single network that models the distribution of training data explicitly to improve the distribution-modeling capability and achieve a higher upper bound of the performance. 

Additionally, although our approach makes use of unlabeled images successfully, a number of normal images are still required for training, which can also be time-consuming to collect in practice. Recently, \citet{zaheer2022generative} proposed the generative cooperative learning (GCL) approach for anomaly detection, which is trained using only unlabeled images where normal samples are the majority. They designed a co-training strategy of an AE and a classifier to generate pseudo labels for unlabeled images. Inspired by this, we intend to explore a more effective pseudo label generation approach with reference to methods of learning with noisy labels \citep{wei2020combating,jiang2018mentornet,han2018co}, to develop a powerful anomaly detection framework without the requirement of any training annotations. 

\subsection{Future directions and challenges}
Considering the current limitations, we summarize several promising emerging directions for anomaly detection: (1) unsupervised anomaly detection \citep{zaheer2022generative} (using only unlabeled images for training to detect anomalies), (2) open-set supervised anomaly detection \citep{ding2022catching} (using a few labeled abnormal images and normal images for training to detect both seen anomalies and unseen anomalies), and (3) few-shot anomaly detection \citep{huang2022registration} (using only a limited number of normal images for training to detect anomalies). Actually, the first step for task (1), handling the unsupervised anomaly detection, is to generate reasonable pseudo labels for unlabeled training images. Once these pseudo normal or abnormal labels for the training data have been obtained, the task (1) can then be decomposed into the two further tasks, tasks (2) and (3).

To explore the three emerging directions, several challenges need to be studied. Firstly, abnormal medical images only have subtle difference to normal ones. This could make it difficult to assign accurate pseudo labels using current methods for learning with noisy labels \citep{wei2020combating}, where predictions are made by vanilla classification networks according to the whole image. Another challenge is that classes of anomalies are inexhaustible. Even if some abnormal images are labeled accurately, incorporating them into training can render models ineffective in generalizing to unseen anomaly classes. In summary, fine-grained models that are able to recognize subtle lesions and a new training paradigm for utilizing limited labeled images are in high demand for anomaly detection.

\section{Conclusion}
\label{sec:Conclusion}
In this paper, we introduce one-class semi-supervised learning (OC-SSL) to utilize known normal and unlabeled data for training, and propose Dual-distribution Discrepancy for Anomaly Detection (DDAD) based on this setting. Two new anomaly scores, intra- and inter-discrepancy, are designed based on DDAD for identifying anomalies. In addition, an Anomaly Score Refinement Net (ASR-Net) trained via self-supervised learning is designed to refine the two anomaly scores. It provides a new perspective on using self-supervised learning to improve anomaly detection and shows better robustness and performance than previous self-supervised methods on various datasets. To facilitate the fair and comprehensive comparison of different methods, we collect and organize five medical datasets that include three modalities and release them as benchmarks for medical anomaly detection. Experiments on the five benchmarks demonstrate that the proposed DDAD with ASR-Net is effective and generalizable, achieving state-of-the-art performance on a wide variety of medical images. Evaluation on unseen diseases further demonstrates the potential of our method for recognition of rare diseases, whose samples are unavailable in the unlabeled data. Results also reveal how to use self-supervised learning for better anomaly detection. Compared with training the network to directly detect anomalies, using indirect strategies, such as applying self-supervised refinement and self-supervised representations, is more promising. As this work presents the first method that utilizes readily available unlabeled images to improve the performance of anomaly detection and provides a comprehensive comparison of various methods on various datasets, we hope it will inspire researchers to explore anomaly detection in a more effective way. We also hope our released benchmarks for medical anomaly detection will facilitate the fair comparison of related works and contribute to the development of this area.

\section*{Acknowledgments}
This work was supported in part by Hong Kong Innovation and Technology Fund (No. ITS/028/21FP), National Natural Science Foundation of China (61872417, 62061160490, 62176098, 61703049) and Natural Science Foundation of Hubei Province of China (2019CFA022).

\bibliographystyle{model2-names.bst}\biboptions{authoryear}
\bibliography{refs}

\end{document}